%% file: main.tex
\documentclass{article}

\PassOptionsToPackage{sort, numbers, compress}{natbib}
\usepackage[preprint]{neurips2024/neurips_2024}

\usepackage[utf8]{inputenc} % allow utf-8 input
\usepackage[T1]{fontenc}    % use 8-bit T1 fonts
\usepackage{hyperref}       % hyperlinks
\usepackage{url}            % simple URL typesetting
\usepackage{booktabs}       % professional-quality tables
\usepackage{amsfonts}       % blackboard math symbols
\usepackage{nicefrac}       % compact symbols for 1/2, etc.
\usepackage{microtype}      % microtypography
\usepackage{xcolor}         % colors

\usepackage{wrapfig}

\usepackage{algorithm}
\usepackage{algorithmic}

\usepackage{amsmath}
\usepackage{amssymb}
\usepackage{mathtools}
\usepackage{xspace}
\usepackage{amsthm}
\mathtoolsset{showonlyrefs} 

\usepackage[capitalize,noabbrev]{cleveref}

\theoremstyle{plain}
\newtheorem{theorem}{Theorem}[section]

\newtheorem{lemma}[theorem]{Lemma}

\theoremstyle{definition}

\newtheorem{assumption}[theorem]{Assumption}
\theoremstyle{remark}

\DeclareMathOperator*{\argsup}{arg\,sup}

\DeclareMathOperator*{\argmax}{arg\,max}

\DeclarePairedDelimiterX{\infdivx}[2]{(}{)}{%
	#1\;\delimsize\|\;#2%
}
\newcommand{\kl}{\textrm{KL}\infdivx}

\newcommand{\ouralgo}{\textsc{BIG}\xspace}

\definecolor{tab_blue}{rgb}{0.0, 0.5, 1.0}
\definecolor{tab_orange}{rgb}{0.93, 0.57, 0.13}
\definecolor{tab_green}{rgb}{0.4, 0.69, 0.2}
\definecolor{darkred}{rgb}{0.8, 0.0, 0.0}
\definecolor{gold}{rgb}{1.0, 0.84, 0.0}

\title{A Bayesian Solution To The Imitation Gap}

\author{%
    Risto Vuorio\thanks{Equal contribution}\hspace{0.4em} \\
    University of Oxford\\
    \texttt{risto.vuorio@cs.ox.ac.uk}
    \And
    Mattie Fellows$^*$\\
    University of Oxford\\
    \And
    Cong Lu$^*$\thanks{Work done while at University of Oxford}\\
    University of British Columbia \\
    Vector Institute \\
    \And
    Clémence Grislain\\
    Sorbonne University \thanks{Work done while interning at University of Oxford}
    \And
    Shimon Whiteson\\
    University of Oxford\\
}

\begin{document}

\maketitle

\input{sections/abstract}
\input{sections/introduction}
\input{sections/preliminaries}
\input{sections/problem_setting}
\input{sections/bayesian_formulation}
\input{sections/experiments}
\input{sections/related}
\input{sections/conclusion}

\clearpage
\bibliography{references}
\bibliographystyle{plainnat}

%%%%%%%%%%%%%%%%%%%%%%%%%%%%%%%%%%%%%%%%%%%%%%%%%%%%%%%%%%%%

\clearpage
\appendix

\input{sections/appendix}

%%%%%%%%%%%%%%%%%%%%%%%%%%%%%%%%%%%%%%%%%%%%%%%%%%%%%%%%%%%%

\end{document}

%% file: sections/abstract.tex
\begin{abstract}
In many real-world settings, an agent must learn to act in environments where no reward signal can be specified, but a set of expert demonstrations is available.
Imitation learning (IL) is a popular framework for learning policies from such demonstrations.
However, in some cases, differences in observability between the expert and the agent can give rise to an \emph{imitation gap} such that the expert's policy is not optimal for the agent and a naive application of IL can fail catastrophically.
In particular, if the expert observes the Markov state and the agent does not, then the expert will not demonstrate the information-gathering behavior needed by the agent but not the expert.
In this paper, we propose a Bayesian solution to the Imitation Gap (\ouralgo), first using the expert demonstrations, together with a prior specifying the cost of exploratory behavior that is not demonstrated, to infer a posterior over rewards with Bayesian inverse reinforcement learning (IRL).
\ouralgo then uses the reward posterior to learn a Bayes-optimal policy.
Our experiments show that BIG, unlike IL, allows the agent to explore at test time when presented with an imitation gap, whilst still learning to behave optimally using expert demonstrations when no such gap exists.
\end{abstract}

%% file: sections/introduction.tex
\section{Introduction}
\label{sec:introduction}
Imitation learning~\citep{schaal1999imitation, hussein2017imitation} is a powerful method for training policies when expert demonstration data is available for the desired behavior, without the need for explicit reward.
However, standard imitation learning algorithms can fail when the expert demonstrator has access to privileged information that the imitator lacks.
Specifically, if the expert observes the full Markov state, but the imitator operates under partial observability, then imitating expert behavior can lead to suboptimal performance~\citep{chen2019lbc, weihs2021bridging,cai2021seeing}.
This mismatch in observability is called the \emph{imitation gap}.

For example, consider training a fruit-picking robot by learning from human demonstrations.
Humans use a combination of visual cues and touch to quickly determine the ripeness of the fruit.
They may reach out for a fruit that looks ripe, but then leave it on the branch to ripen more if it feels too hard in their hand.
However, the robot has to rely on the visual cues alone since it lacks the sense of touch.
If it then picks all fruit that look ripe on one side, it may end up picking many underripe fruit.
A more intelligent robot would turn around fruit that looks borderline ripe to visually inspect them from all sides without detaching them from the branch.
This exploratory behavior is never demonstrated, leading to an imitation gap, and hence na\"ively imitating expert demonstrations leads to suboptimal behavior.

At first glance, this problem seems intractable, since we cannot imitate behavior that is not demonstrated.
Indeed, many prior solutions to the imitation gap often require privileged access to online reward information.
The key insight in this paper is that the remaining uncertainty may be characterized in the form of a prior, specified over the cost of exploration in unobserved states.
This leads us to propose a fully \textbf{B}ayesian solution to the \textbf{I}mitation \textbf{G}ap (\ouralgo) which learns to behave Bayes-optimally at test-time, although its value will naturally be a function of the agent's uncertainty.

We demonstrate how the prior can integrate several sources of information available before test time: we specify an initial reward prior, from which we may infer a posterior given a set of expert demonstrations and simulator, using Bayesian inverse reinforcement learning \citep[BIRL]{Ramachandran07} with successor features~\citep{Barreto17,Brown2019,filos2021psiphilearning}.
We also specify a reward prior over key exploration states to specify uncertainty about the \emph{cost of exploration} where there is an imitation gap.
This allows \ouralgo to optimally trade off any remaining uncertainty about the true environment state using a Bayes-optimal policy, with the predictive reward under these priors as the fixed belief over rewards at test time.
When there is an imitation gap, the reward prior can influence the agent's behavior instead, yielding policies that \emph{balance exploration to reduce the uncertainty in the environment with exploitation of expert demonstrations}.
When there is no imitation gap, the agent can directly imitate expert behavior.

We evaluate \ouralgo across a number of standard imitation gap problems, and further show that it scales to environments with high-dimensional observations.
In each case, we can recover suitable reward functions from \emph{only expert demonstrations and the cost of exploration prior}.

%% file: sections/preliminaries.tex
\section{Preliminaries}
\textbf{Reinforcement Learning.}
We model the environment as a Markov Decision Process~\citep[MDP]{Sutton1998}, defined as a tuple $\mathcal{M}\coloneqq\langle\mathcal{S}, \mathcal{A}, p(s_{t+1}|s_t, a_t), p(s_0), r(s_t, a_t),\gamma\rangle$, where $\mathcal{S}$ and $\mathcal{A}$ denote the state and action spaces respectively, $p(s_{t+1}|s_t, a_t)$  the transition dynamics, $r(s_t, a_t)$  the reward function, and $p(s_0)$ the initial state distribution.
We denote a sampled reward $r$.
The goal in reinforcement learning is to optimize a policy $\pi (a | s)$ that maximizes the expected return $\mathbb{E}_{\pi, p}\left[\sum_{t=0}^\infty \gamma^t r(s_t, a_t)\right]$.

\textbf{Learning from Demonstrations.}
We assume access to a set of expert demonstrations $\mathcal{D}_\textrm{Expert}\coloneqq \{\tau_i\}_{i=1}^{N_\textrm{Expert}}$ of state-action trajectories $\tau_i\coloneqq \{s_0,a_0,s_1,a_1,\dots \}$.
In inverse reinforcement learning (IRL), the objective is to use these demonstrations to learn the reward function that the expert maximizes.
In typical IRL methods, such as maximum entropy IRL~\citep{ziebart2008maximum}, the reward function is learned alongside a policy that optimizes that function in a bi-level optimization algorithm.

By contrast, imitation learning is a closely related approach that seeks to directly mimic expert behavior from the demonstrations.
This can be expressed as matching the distribution of the state-action pairs generated by the imitator to that of the expert~\citep{ho2016generative,finn2016guided}, i.e., minimizing the divergence between the limiting distribution over state-action pairs $D(\rho^\ast(s,a) \| \rho_{\pi}(s,a))$ where $\rho^\ast$ and $\rho_\pi$ are the state-action marginal distributions of the expert and the imitator policies respectively.

\textbf{Successor Features.}
Successor features~\citep[SFs]{Barreto17} are a value function representation that decouples the dynamics of the environment from the reward. Suppose that the reward function of the environment could be computed linearly as $r(s, a) = \nu(s,a)^\top \omega$ where $\nu(s,a)\in\mathbb{R}^d$ are features and $\omega\in\mathbb{R}^d$ are weights.
For any given policy, we may then factor the $Q$-function as $Q^\pi(s, a,\omega)=\Psi^\pi(s,a)^\top \omega$
where $\Psi^\pi(s,a) :=\mathbb{E}_{\tau\sim p^{\pi}} [\sum_{t=0}^\infty\gamma^t\nu(s_t,a_t)\vert s_0=s,a_0=a]$ is the successor feature of $(s, a)$ under $\pi$.

%% file: sections/problem_setting.tex
\section{Problem Setting}
\label{sec:problem_setting}
We now formalize the imitation gap in terms of a contextual MDP (CMDP) where the demonstrator, but not the imitator, observes a hidden parameter $\theta\in\Theta$ that affects the environment dynamics.
Whilst the imitation gap could also be formalized in terms of partial observability in the state, we choose the CMDP formulation to make explicit what is hidden from the imitator and assume the states are fully observable.
Formally, we define a CMDP as:
\begin{align}
\small
\mathcal{M}(\theta)\coloneqq\langle\mathcal{S}, \mathcal{A}, p(s_{t+1}|s_t, a_t, \theta), p(s_0), r(s_t, a_t),\gamma\rangle, \label{eqn:contextual_mdp}
\end{align} 
with an underlying distribution over contexts $p(\theta)$.
The reward function is independent of $\theta$ as we assume all tasks in the CMDP have a common goal (for example, driving safely at a junction), but differ in their state transition dynamics.
We are interested in policies that optimize the contextual expected, discounted return $\mathbb{E}_{p^{\pi}_\infty ( \theta)}\left[\sum_{t=0}^\infty \gamma^t r(s_t,a_t)\right]$ where $p^{\pi}_\infty(\theta)$ is the distribution over infinite-horizon trajectories associated with policy $\pi$ and context $\theta$.

At test time, we are interested in behaving optimally in a CMDP $\mathcal{M}(\theta_\textrm{test})$ allocated according to the prior $\theta_\textrm{test}\sim p(\theta)$. Like other successor feature-based approaches \citep{Barreto17,Brown2019,Janz19,filos2021psiphilearning} we make the following assumption about the reward parametrization:
\begin{assumption}\label{ass:reward}
The underlying reward function is bounded and can be represented as a linear function with respect to a reward feature vector $\nu(s,a)$ that is known a priori such that $r(s,a;\omega^\star)=\nu(s,a)^\top \omega^\star\in[r_\textrm{min},r_\textrm{max}]$.
Furthermore, the reward function is shared across all CMDPs, i.e., independent of $\theta$.
\end{assumption}

This assumption simplifies our analysis and enables the derivation of convex optimization procedures.
We do not assume oracle access to perfect reward features.
Instead, the features could be the result of an unsupervised learning step external to our algorithm.
Crucially, the agent can observe states and choose actions according to a policy but does \emph{not} observe rewards nor $\theta_\textrm{test}$ and does not know the true reward parametrization $\omega^\star$. 
Instead, the agent is given a dataset of $N_\textrm{Expert}$ demonstrations $\mathcal{D}_\textrm{Expert}\coloneqq \{\tau_i\}_{i=1}^{N_\textrm{Expert}}$ of state-action trajectories $\tau_i\coloneqq \{s_0,a_0,s_1,a_1,\dots \}$ of length $H_i$ in CMDPs sampled from $p(\theta)$, where each expert $\pi^\star_\textrm{Expert}(\cdot,\theta_i)$ behaves optimally in its assigned CMDP, e.g., expert trajectories of a car turning at a junction in summer and winter.
Furthermore, the agent has access to a simulator, where CMDPs are allocated according to the prior $p(\theta)$, and the agent can interact with the corresponding environment, observing a trajectory of state-action pairs.
We denote the complete dataset of simulated trajectories as $\mathcal{D}_\textrm{Simulator}\coloneqq \{\tau_i\}_{i=1}^{N_\textrm{Simulator}}$.
The agent never directly observes rewards nor $\theta_i$ in either $\mathcal{D}_\textrm{Expert}$ or $\mathcal{D}_\textrm{Simulator}$.

At test time, the agent is assigned a CMDP according to $\theta_{\text{test}} \sim p(
\theta)$ and interacts with $\mathcal{M}(\theta_{\text{test}})$, obtaining a history of state-actions: $h_t\coloneqq\{s_0,a_0,\cdots a_{t-1},s_t\}$ at time $t$ and takes actions according to a history-conditioned policy: $s_t\sim \pi(h_t)$.
The agent never observes the reward history. 

\subsection{The Tiger-Treasure Problem}
\label{sec:counterexample}
\begin{wrapfigure}{r}{0.4\textwidth}
\centering
\vspace{-8mm}
\includegraphics[width=0.38\textwidth]{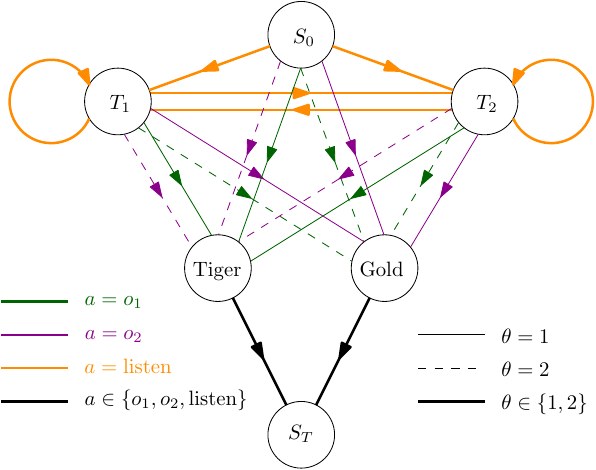}
% \vspace{-2mm}
\caption{
\small
A diagram of the Tiger-Treasure Problem MDP, a classic example of an imitation gap. The agent initially does not know which door the treasure or tiger is behind and must take listening actions to resolve its uncertainty.}
\vspace{-3mm}
\label{fig:tiger_treasure}
\end{wrapfigure}
We introduce a variant of the classic ``Tiger-Treasure Problem'' from \citet{kaelbling1998planning} to illustrate how na\"{i}vely applying imitation learning fails when there is an imitation gap.
Consider the CMDP in \Cref{fig:tiger_treasure} indexed by $\theta\in\{1,2\}$, representing which door the tiger is behind.
The prior is $p(\theta=1)=p(\theta=2)=0.5$.
In both CMDPs, the agent starts in state $S_0$ and the set of actions available is $\mathcal{A}=\{o_1,o_2,\textrm{listen}\}$, where $o_1$ and $o_2$ open the corresponding door and `listen' listens for a tiger.
In any state $s\in \{S_0,T_1,T_2\}$, the agent transitions deterministically to state $\textrm{Tiger}$ if $a=o_1$ and $\theta=1$ or $a=o_2$ and $\theta=2$, and conversely for the $\textrm{Gold}$ state.
The goal of the agent is to reach the treasure (labeled `$\textrm{Gold}$') whilst avoiding the $\textrm{Tiger}$.
The agent receives a reward $r(\textrm{Gold},\cdot)=10$ for finding the gold and $r(\textrm{Tiger},\cdot)=-100$ for finding the tiger.
After this, the agent transitions to terminal state $S_T$ regardless of action taken.

The agent can also listen before any doors are opened, receiving a stochastic signal with success rate $p>0.5$ correlated with the identity of the door with the tiger.
Hearing the tiger in room $i$ transitions the agent to state $T_i$.
For $s\in \{S_0,T_1,T_2\}$ and $a=\textrm{listen}$, the agent transitions to state $T_1$ with probability $p$ if $\theta=1$ and $T_2$ with $p$ if $\theta=2$, otherwise it transitions to state $T_2$ with probability $1-p$ if $\theta=1$ and $T_1$ with $1-p$ if $\theta=2$.
States $T_1$ and $T_2$ are not present in \citet{kaelbling1998planning}'s original problem, but are included here because reward is assumed independent of $\theta$ and hence partial observability about the MDP is encoded in the state. Entering a listening state incurs a small negative reward $r(T_1,\cdot )=r(T_2,\cdot)=-1$.
All other rewards not specified are $0$.

For ease of exposition, assume the listening success $p=1$.
The expert has privileged knowledge about the MDP, and always chooses to open the door with the gold behind it.
Consequently, expert demonstrations never feature listening actions.
At test time, there is an imitation gap, as the agent does not know a priori which door the tiger is behind.
A na\"ive imitator does not realize that the expert is conditioning on extra information and so thinks the expert is randomly choosing which door to open; imitating that gives suboptimal return of $-45\gamma$.
By contrast, an agent that chooses to listen always receives a return of $10\gamma-1$ if it acts optimally on the revealed location of the tiger.
This example demonstrates the failure of na\"ive imitation learning in simple settings when there is an imitation gap. 

%% file: sections/bayesian_formulation.tex
\section{Towards a Bayesian Solution}
\label{sec:full_Bayes}

\begin{wrapfigure}{r}{0.5\textwidth}
\centering
\includegraphics[width=0.48\textwidth]{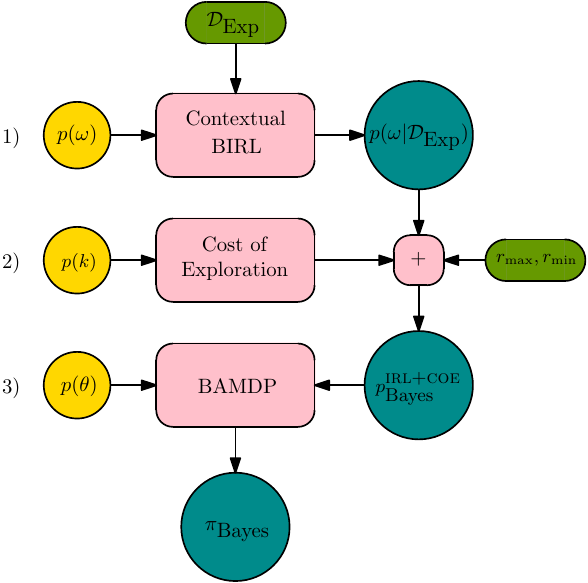}
\caption{
\small
Schematic of the Bayesian solution to the Imitation Gap (\ouralgo).
Prior information is shown in \textcolor{tab_green}{\textbf{green}}, algorithms in \textcolor{pink}{\textbf{pink}}, prior distributions in \textcolor{gold}{\textbf{yellow}}, and outputs in \textcolor{tab_blue}{\textbf{blue}}. 
}
\label{fig:schematic}
\vspace{-6mm}
\end{wrapfigure}

In this paper, we propose a \textbf{B}ayesian solution to the \textbf{I}mitation \textbf{G}ap (\ouralgo), where the goal is to learn a policy that can optimally trade-off its uncertainty at test time with imitating the expert demonstrator.
Unlike in the canonical Bayesian RL setting, our formulation does not have access to reward samples from which to infer the underlying reward function.
Instead, the reward prior determines the cost of exploration (COE) for the agent at test time, meaning that the agent can still behave optimally under partial observability.
We provide an overall schematic of our approach in \Cref{fig:schematic}. 

\ouralgo has three main phases which are labeled in \Cref{fig:schematic}.
In the first, we provide an initial prior over the reward parametrization $p(\omega)$.
We integrate information from the expert data $\mathcal{D}_\textrm{Exp}$ into the prior using a novel \emph{contextual} Bayesian IRL (BIRL) approach (\Cref{sec:contextual_successor_features,sec:inferring_reward}) that infers the posterior $p(\omega\vert \mathcal{D}_\textrm{Exp})$.
Because entire classes of reward functions can explain expert data equally well, IRL is an underspecified problem and approaches can equally penalize any unvisited state.

In the second phase (\cref{sec:coe_prior}), we restrict the class of reward functions to those that allow for exploration at test time, by first normalizing the posterior so that the predictive reward lies in the range $[r_\textrm{min}, r_\textrm{max}]$ according to \Cref{ass:reward}.
Representing reward as $r=k\times r_\textrm{max}$, we specify a cost of exploration prior $p(k)$ over rewards at key state-action pairs unseen in the expert demonstrations to integrate the cost of exploration information at test time.
This approach ensures that all rewards still belong to the same \emph{class} \citep[Definition 1]{Rafailov23} after the IRL stage, i.e., yielding an equivalent optimal policy.
This step is required as the expert demonstrations do not have full coverage. $p(k)$ encodes the relative cost of deviating from an optimal policy which encourages exploration for the downstream policy, complementing the IRL data.
Integrating both the IRL and COE priors, we denote the Bayesian reward distribution as $p^\textrm{IRL+COE}_\textrm{Bayes}(r\vert s,a)$.
We present pseudocode for an algorithm implementing the first and second phase in \cref{app:main_algorithm}.

In the third phase, we solve a Bayesian RL problem in which the goal is to learn a Bayes-optimal policy $\pi_\textrm{Bayes}$ that can be deployed at test time.
As shown in \Cref{fig:schematic}, the inputs to the Bayes-adaptive MDP (BAMDP) are the Bayesian reward distribution $p^\textrm{IRL+COE}_\textrm{Bayes}(r\vert s,a)$ and a prior over context variables $p(\theta)$ (\Cref{sec:BAMDP_definition}).
The agent extracts reward information from the expert trajectories to learn a predictive reward but does not directly imitate the expert's behavior and thus can adapt to the unknown MDP at test time, avoiding the problems with na\"ive imitation learning discussed in \Cref{sec:counterexample}.
Due to the diverse nature of the sources of input information, inferring the reward posterior requires the agent to learn and maintain several distributions over random variables, we summarize them in \Cref{table:distributions} in \Cref{appsec:dist_summary}.

\subsection{Contextual Successor Features}
\label{sec:contextual_successor_features}
Before performing BIRL, we must learn a contextual value function representation to define the likelihood over expert trajectories.
For an agent following policy $\pi$, we can characterize the expected discounted return as a function of state-action pairs via the contextual $Q$-function $Q^\pi(s,a,\theta,\omega)$, which satisfies the contextual Bellman equation: $\mathcal{B}^\pi[Q^\pi](s,a,\theta,\omega)=Q^\pi(s,a,\theta,\omega)$ where $\mathcal{B}^\pi[Q^\pi](s,a,\theta,\omega)$ is the contextual Bellman operator: $\mathcal{B}^\pi[Q^\pi](s,a,\theta,\omega)\coloneqq \nu(s,a)^\top\omega+\gamma \mathbb{E}_{s'\sim p(s'\vert s,a,\theta)a'\sim \pi(a'\vert s')}\left[Q^\pi(s',a',\theta,\omega)\right].$

\Cref{ass:reward} specifies a linear reward function,
which  allows us to use a \emph{successor feature} representation of the $Q$-function that factors the reward parametrization out of $Q^\pi(s',a',\theta,\omega)=\Psi^\pi(s',a',\theta)^\top \omega$
where $\Psi^\pi(s,a,\theta)$ is the contextual successor feature, defined as: $\Psi^\pi(s,a,\theta)=\mathbb{E}_{\tau\sim p^{\pi} (\tau_\infty\vert \theta)}\left[\sum_{t=0}^\infty\gamma^t\nu(s_t,a_t)\bigg\vert s_0=s,a_0=a\right]$.
N.b., the reward is linear w.r.t. the features $\nu(s, a)$, which could themselves be learned and arbitrarily complex.
Learning $\Psi^\pi(s,a,\theta)$ means that we do not need to solve a Bellman equation every time $\omega$ changes; we can simply take a dot product between the existing successor feature and the new $\omega$.
\cref{app:bsfl} details the training process.

\subsection{Contextual Bayesian IRL}
\label{sec:inferring_reward}
We now describe the first phase of our pipeline in \Cref{fig:schematic}.
Inferring the reward function is a Bayesian regression problem.
As is typical for regression problems \citep{Murphy2013}, we specify a Gaussian reward model $p(r\vert s,a,\omega)=\mathcal{N}(\nu(s,a)^\top\omega,I\sigma^2)$ with mean $\nu(s,a)$ and scalar variance parameter $\sigma\in\mathbb{R}$. 
We specify a Gaussian prior over the unknown reward parameterization $\mathcal{N}(\omega\vert \omega_0, I \sigma_0^2)$ where
$\omega_0$ is the prior mean and prior variance $\sigma_0^2>0$ represents the belief in $\omega_0$.
Tuning $\sigma_0^2$ thus allows us to set how much the expert's trajectories affect the prior reward specification after BIRL.
We now exploit expert data to refine our prior by leveraging approaches from BIRL.

To derive the likelihood, it is necessary to define a model of the expert observations.
In the classic Bayesian IRL approach \citep{Ramachandran07}, a likelihood is specified for a single MDP: $p(\tau_i\vert \omega)$.
To adapt our approach to experts that act in multiple MDPs, our likelihood should account for the context $p(\tau_i\vert \theta_i,\omega)=\prod_{j=0}^{T_i-1} p(s_j,a_j\vert \theta_i,\omega)$. 
We assume that the agent's policy is represented by a potential function defined by the optimal $Q$-function for the agent's MDP:
\begin{align}
% \small
p( a\vert s, \theta_i,\omega)=\frac{1}{z(s,\omega,\theta_i)}\exp\left(\frac{1}{\alpha}\Psi(s,a,\theta_i)^\top \omega\right),\label{eq:boltzman_likelihood}
\end{align}
where $z(s,\omega,\theta)$ is the normalization constant. Here $\alpha$ is a temperature parameter that controls how optimal the expert's actions are with respect to $Q^\star(s,a,\theta_i,\omega)=\Psi^\star(s,a,\theta_i)^\top \omega$. As is convention \citep{Ramachandran07,Arora2021}, this assumption stabilizes inference algorithms by softening the Dirac delta policy that the agent is following, allowing for gradients to flow into the density.
A deterministic optimal policy is recovered in the limit  $\alpha\rightarrow 0$. 

Given the likelihood and prior, we infer the expert reward posterior $p(\omega\vert \mathcal{D}_\textrm{Expert})$. Marginalizing, we obtain the Bayesian reward distribution:
\begin{align}
% \small
p^\textrm{IRL}_\textrm{Bayes}(r\vert s,a)\coloneqq \mathbb{E}_{\omega \sim p(\omega\vert  \mathcal{D}_\textrm{Expert})}\left[ p(r\vert s,a,\omega)\right],
\end{align}
which incorporates both the epistemic uncertainty from the posterior and the aleatoric uncertainty from the reward model. Analogously to Bayesian logistic regression~\citep{Murphy2013}, using the potential function from \Cref{eq:boltzman_likelihood} does not yield a closed-form solution for the posterior. Instead, we use the Laplace approximation for the posterior:
\begin{theorem} \label{proof:laplace_parameter} Define $\varsigma_0^2 \coloneqq \frac{\sigma_0^2}{\alpha}$. Using the Laplace approximation, the approximate posterior is $p(\omega\vert \mathcal{D}_\textrm{Expert})\approx \mathcal{N}(\omega\vert \omega^\star_\textrm{Laplace},\Sigma_\textrm{Laplace})$ where $\Sigma_\textrm{Laplace}=\nabla^2_\omega \log p(\omega^\star_\textrm{Laplace}\vert \mathcal{D}_\textrm{Expert})$ and  $\omega^\star_\textrm{Laplace}$ is the MAP estimate, which can be found  by carrying out the following stochastic gradient descent updates on the log-posterior:
\begin{align}
% \small
\omega \leftarrow &\omega + \eta_\omega \Big( N_\textrm{Expert}{H_i} \bigg(\Psi^\star(s_j,a_j,\theta_i)-\mathbb{E}_{a_i\sim p(a_i\vert s_j,\omega,\theta_i)}\left[ \Psi^\star(s_j,a_i,\theta_i) \right]\Big)-\frac{(\omega-\omega_0)}{\varsigma_0^2}\bigg).\label{eq:stochastic_updates}
\end{align} 
% \vspace{-1cm}
\begin{proof}
    See \Cref{app:proofs}.
\end{proof}
\end{theorem}
The Bernstein von-Mises theorem formally justifies the approximation \citep{Doob49,van1998}, proving that under mild regularity assumptions, the posterior tends to the Laplace approximation in the limit of increasing data. $\varsigma_0^2$ controls how the prior influences learning; as $\varsigma_0^2\rightarrow 0$, expert data is ignored and $\omega^\star_\textrm{Laplace}=\omega_0$.  The posteriors over each expert's contextual variable $p(\theta_i\vert \omega,\tau_i)$ typically have no closed-form analytic solution. We can approximate each $p(\theta_i\vert \omega,\tau_i)$ using variational inference instead, as detailed in \Cref{app:approximate_inference}.

\textbf{Role of Temperature.}
As IRL is underspecified, many reward functions can explain the expert data.
A key insight from \Cref{proof:laplace_parameter} is the role of the temperature parameter $\alpha$ in determining the relative difference between the lowest and highest rewards assigned.
The example in \Cref{app:balancing_prior} illustrates that when $\alpha\rightarrow0$, the expert policy model becomes more deterministic, always choosing the action with the highest return.
Arbitrarily small differences between rewards explain expert behavior, so IRL pulls the parametrization difference to be as small as possible. Conversely, when $\alpha\rightarrow \infty$, the expert policy becomes more stochastic, taking low-return actions more frequently in proportion to their value.
An increasingly large separation in rewards is needed to explain expert behavior. 
% \end{remark}

\subsection{Cost of Exploration Prior}
\label{sec:coe_prior}

For the second phase, we specify a prior over the reward to incorporate information on the cost of exploration into the posterior reward returned by the contextual BIRL to refine the solution.
Note that the imitation gap problem would not be solvable without access to prior information defining the cost of exploration, as the expert data does not demonstrate which states are safe to explore and how to balance exploration and exploitation.
This prior is only introduced for exploration and does not need to contain any information about the exploitation, as that is learned from the expert demonstrations.
Furthermore, the prior could be arbitrarily uninformative, in the third phase, our method learns the Bayes-optimal policy for any prior.

As shown in \Cref{fig:schematic}, we start by rescaling the posterior reward to be within the bounds $[r_\textrm{min}, r_\textrm{max}]$ according to \Cref{ass:reward}.
We denote the corresponding scaled reward parametrization as $\bar{\omega}^\star_\textrm{Bayes}$.
Given the infinite horizon, any linear transformation applied to the reward function results in the same optimal policy.

We assume that we know a subset of state-action pairs denoted as $\mathcal{S}_\text{COE} \times \mathcal{A}_\text{COE}$ where exploration can be performed.
We illustrate this for the Tiger-Treasure problem from \Cref{fig:tiger_treasure}; as rewards only depend on the state in this problem, the set is $\mathcal{S}_\text{COE}=\{T_1, T_2\}$.
We introduce a simple COE inside the rescaled IRL reward parameterized by $\bar\omega^\star_\textrm{Bayes}$ by specifying a reward function $p(r\vert s,a,k) = \mathcal{N}(r\vert kr_\textrm{max},\sigma^2)$ over $(s,a)\in\mathcal{S}_\text{COE}\times \mathcal{A}_\text{COE}$ for some $k\in [\frac{r_\textrm{min}}{r_\textrm{max}},1)$; this ensures the mean is contained in $[r_\textrm{min},r_\textrm{max}]$.
A scale $k$ that varies across action-state pairs may also be specified if a more complex cost of exploration information needs to be modeled.
In the simple Tiger-Treasure problem, we know from construction the set of states where exploration can be performed.
In a practical setting with large state space, these states could be obtained by considering non-expert, but safe behaviors from other policies acting in the same environment.
This could require density estimation for continuous state spaces.
We leave development of task specific cost of exploration priors for future work and focus on demonstrating the general principles in this work.

The value of $k$ determines how risk-averse the agent is at test time.
For $k\approx 1$ the agent values exploratory state-actions in $\mathcal{S}_\text{COE}\times \mathcal{A}_\text{COE}$ nearly as much as the most rewarding state-actions learned from IRL.
As such, the cost of exploration is low, and the agent explores until it is highly certain about avoiding low reward actions in the imitation gap, encouraging conservative behavior.
Conversely, as $k\rightarrow\frac{r_\textrm{min}}{r_\textrm{max}} $ the agent becomes less risk-averse and recovers a purely behavioral cloning regime, taking actions that could lead to low reward as they have similar value to exploratory actions.

To incorporate epistemic uncertainty in $k$, we specify a prior $p(k)$ with support over $[\frac{r_\textrm{min}}{r_\textrm{max}},1)$.
Marginalizing, we obtain the Bayesian reward distribution over $\mathcal{S}_\text{COE}\times \mathcal{A}_\text{COE}$:
\begin{align}
% \small
p^{\textrm{COE}}_\textrm{Bayes}(r\vert s,a)=\mathbb{E}_{k\sim p(k)}\left[p(r\vert s,a,k)\right].
\end{align}
For all other state-action pairs, the Bayesian reward distribution remains unchanged, yielding the distribution over $\mathcal{S}\times\mathcal{A}$:
\begin{align}
% \small
p_\textrm{Bayes}^\text{IRL + COE}(r\vert s,a)=
\begin{cases}
  p^{\textrm{COE}}_\textrm{Bayes}(r\vert s,a)& s,a \in \mathcal{S}_\text{COE} \times \mathcal{A}_\text{COE}, \\
p^{\textrm{IRL}}_\textrm{Bayes}(r\vert s,a)& \text{otherwise.}\label{eq:bayes_reward}
\end{cases}
\end{align}

\subsection{Bayes-Optimal Policy Learning}
\label{sec:BAMDP_definition}
For the third phase, we perform Bayesian reinforcement learning, which optimally trades off exploration and exploitation by conditioning actions on the agent's uncertainty over $\theta$.
We can define a Bayes-adaptive MDP~\citep[BAMDP]{duff2002optimal} using the contextual MDP in \Cref{eqn:contextual_mdp} as a model.
At test time, the agent is assigned an MDP $\theta_\textrm{test}\sim p(\theta_\textrm{test})$ and can observe samples from $p(s'\vert s,a,\theta_\textrm{test})$ by interacting with the MDP via its policy.
A history of interactions is denoted $h_t \coloneqq\{s_0,a_0,s_1,a_1,\ldots s_t\} \in \mathcal{H}_t$ where $\mathcal{H}_t$ is the corresponding state-action product space.
After observing a history $h_t$ from the assigned MDP, the agent updates its belief in $\theta_\textrm{test}$ according to the posterior: 
\begin{align}
% \small
p(\theta_\textrm{test}\vert h_t)=\frac{\prod_{i=1}^{t} p(s_t\vert s_{t-1},a_{t-1},\theta_\textrm{test})p(\theta_\textrm{test})}{\mathbb{E}_{\theta_\textrm{test}\sim p(\theta_\textrm{test})}\left[\prod_{i=1}^{t} p(s_t\vert s_{t-1},a_{t-1},\theta_\textrm{test}) \right]}.
\end{align}
Using the posterior, we can define the Bayesian transition distribution as: 
\begin{align}
% \small
p(s_{t+1}\vert h_t,a_t)=\int_\Theta p(s_{t+1}\vert s_t,a_t,\theta_\textrm{test}) p(\theta_\textrm{test}\vert h_t) d\theta_\textrm{test}.
\end{align}
As there is no reward signal available to the agent and rewards do not depend on $\theta$, the Bayesian reward distribution in the BAMDP is exactly the combined Bayesian reward distribution from \Cref{eq:bayes_reward}: $p(r_t\vert h_t,a_t)=p_\textrm{Bayes}^\text{IRL + COE}(r_t\vert s_t,a_t)$. We denote the joint reward-state Bayesian transition distribution as $p(r_t,s_{t+1}\vert s_t,a_t )=p(r_t\vert h_t,a_t)p(s_{t+1}\vert h_t,a_t)$, which is equivalent to the predictive trajectory transition distribution: $p(\tau_{t+1}\vert \tau_t,a_t)=p( \tau_t,a_t,r_t,s_{t+1}\vert \tau_t,a_t)=p(r_t,s_{t+1}\vert h_t,a_t)\underbrace{p(\tau_t,a_t\vert \tau_t,a_t)}_{=1}=p(r_t,s_{t+1}\vert h_t,a_t)$.
Here $p(\tau_{t+1}\vert \tau_t,a_t)$ is used to reason over unobserved counterfactual trajectories, and so must account for predictive reward. 
We define the corresponding BAMDP similarly to \citet{fellows2023bayesian} as
$\mathcal{M}_\textrm{BAMDP}\coloneqq \langle\mathcal{T}, \mathcal{A}, p(\tau_{t+1}\vert \tau_t,a_t), p(s_0), \gamma \rangle$ where $\mathcal{T}$ is the space of all possible trajectories.

In Bayesian RL, policies $\pi_\textrm{Bayes}(a_t\vert h_t)$ map from \emph{histories} to distributions over actions, and our goal is to learn a Bayes-optimal policy $ \pi_\textrm{Bayes}^\star(a_t\vert h_t) $ that solves $\mathcal{M}_\textrm{BAMDP}$.
Due to the linearity of our formulation, we show in \cref{app:BRL_objective_derivation} that solving the BAMDP is equivalent to optimizing the following Bayesian RL objective for $\pi^\textrm{Bayes}$:
\begin{align}
% \small
J^\pi_\textrm{Bayes}=\mathbb{E}_{p(\theta_\textrm{test})}\left[ \mathbb{E}_{p(h_\infty\vert \theta_\textrm{test})}\left[ \sum_{i=0}^\infty \gamma^i r_\textrm{Bayes}^\text{IRL + COE}(s_i,a_i) \right] \right],
\end{align}
where $p(h_\infty\vert \theta_\textrm{test})=p_0(s_0)\prod_{i=0}^\infty p(s_{i+1}\vert s_i,a_i,\theta_\textrm{test})\pi(a_i\vert h_i)$, and $r_\textrm{Bayes}^\text{IRL + COE}(s,a)$ is the predictive reward:
\begin{align}
% \small
r_\textrm{Bayes}^\textrm{IRL + COE}(s,a)\coloneqq
\begin{cases}
r_\textrm{max} \mathbb{E}_{k\sim p(k)}\left[k\right]& s,a \in \mathcal{S}_\text{COE} \times \mathcal{A}_\text{COE}, \\
\nu^\top(s,a)\bar{\omega}^\star_\textrm{Bayes} & \text{otherwise.}\label{eq:predictive_reward}
\end{cases}
\end{align}

In the following empirical evaluation, we approximate Bayes-optimal policies by training a policy conditioned on the true inference model using DQN~\citep{mnih2013playing}.

%% file: sections/experiments.tex
\section{Empirical Evaluation}
\label{sec:experiments}
To evaluate \ouralgo, we conduct experiments across a series of imitation gap problems.
In all the tested environments, the agent has to solve a task that requires exploration.
First, we demonstrate that in the Tiger-Treasure environment, naive IRL learns a reward function that does not lead to the desired exploratory policy, whereas \ouralgo does by exploiting prior information about the cost of exploration.
Second, we illustrate that, by marginalizing over the context distribution, naive IRL can learn a reward function that does not capture the expert's intent.
Finally, we present results in a gridworld environment to show that \ouralgo can handle imitation learning tasks with larger state-action spaces.
Since we assume no access to true environment reward or expert state information, most previous solutions to the imitation gap are not applicable.
Instead, we compare to maximum entropy IRL (labeled `No-Prior' in our experiments).
For convenience, we implement \Cref{alg:big} using the true posterior $p(\theta \vert \tau)$ as the inference model.
In all experiments, we use deep neural networks for $\Psi$.
We assume normally distributed errors and report standard error across seeds in the figures.

\subsection{Investigating Reward Priors}
\begin{wrapfigure}{r}{0.5\textwidth}
\centering
\vspace{-3mm}
\includegraphics[width=0.245\textwidth]{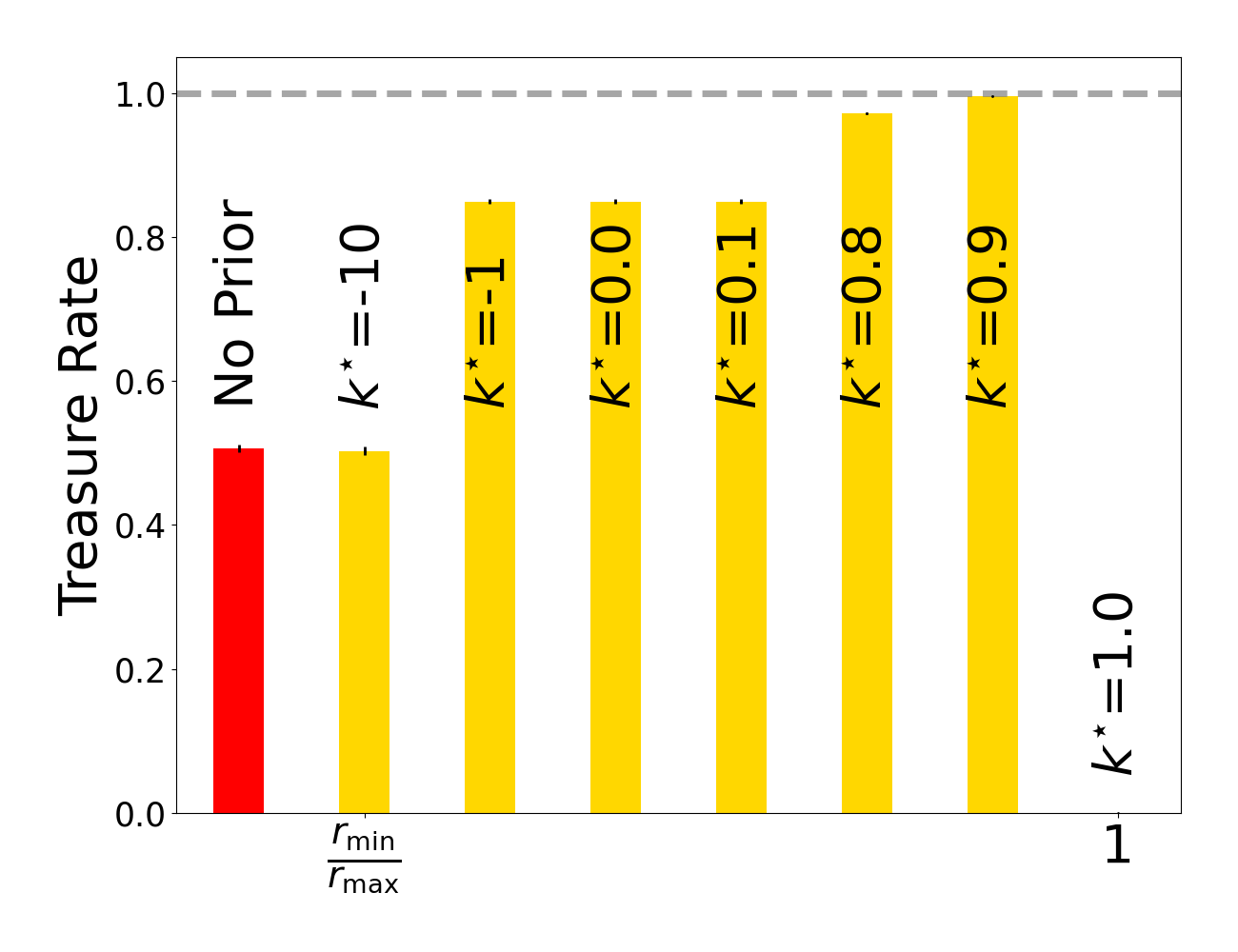}
\includegraphics[width=0.245\textwidth]{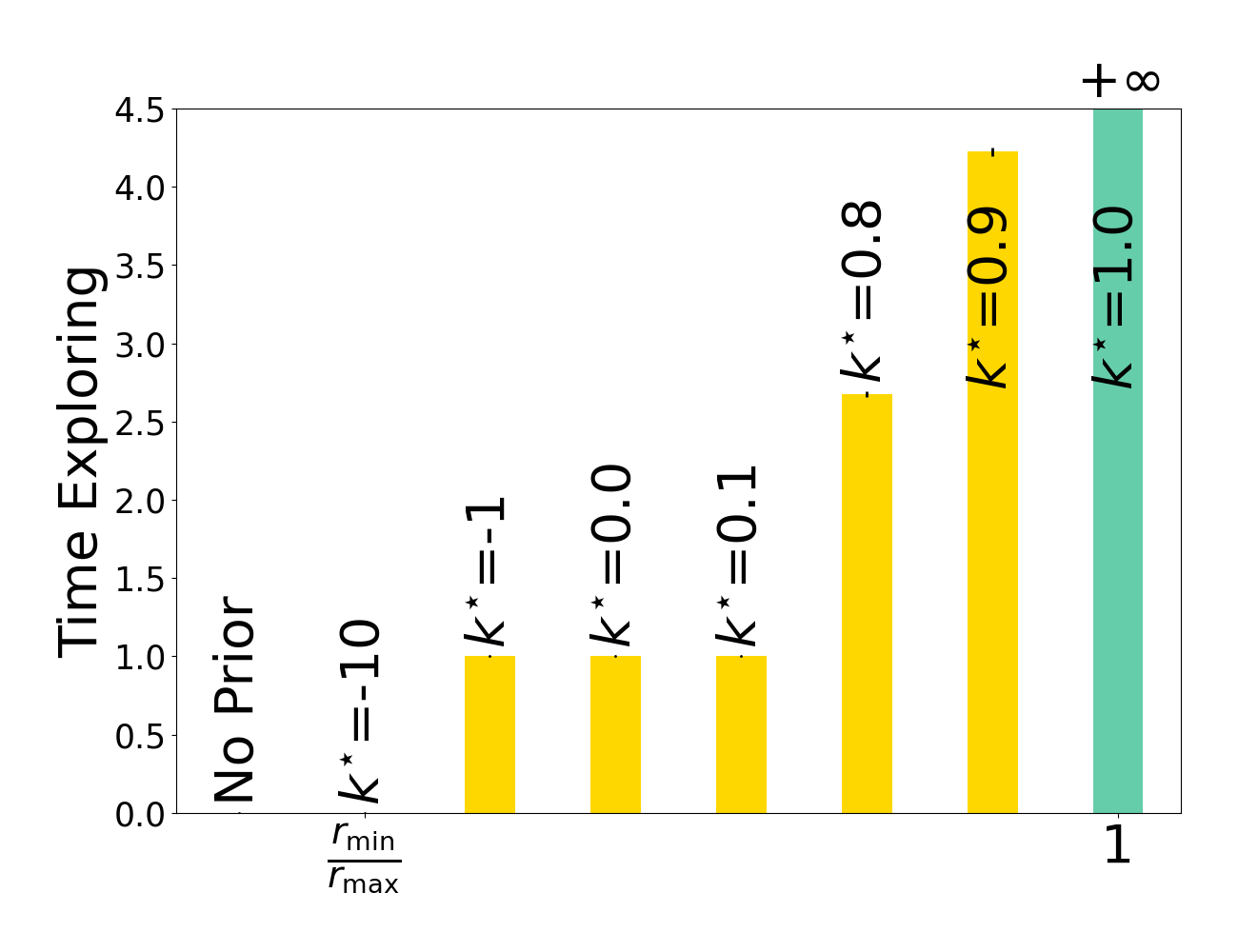}
\vspace{-5mm}
\caption{\small
Evaluation of \ouralgo in the Tiger-Treasure environment.
Success rate and time exploring (in steps) for policies learned with a uniform prior reward with different means are represented in yellow ($k^\star < 1$) and in green ($k^\star=1$), while the case with no prior is shown in red. Error bars indicate the standard error of the mean across 10 seeds.
The symbol $+\infty$ indicates that, for some trials, the agent explores throughout the entire (infinite) episode.
}
\vspace{-2mm}
\label{fig:tt_scaling_constant}
\end{wrapfigure}
\label{subsec:tiger-treasure}
In the Tiger-Treasure environment introduced in \Cref{sec:counterexample}, since the expert always goes to the treasure room directly, we cannot extract information about optimal exploration from the expert data.
As discussed in \Cref{sec:coe_prior}, we use the prior $p(k)$ to enable exploratory behavior at test time.
We explore the influence of this reward prior on the environment from \Cref{fig:tiger_treasure}, using a space of uniform priors $p(k)=\textrm{Unif}\left([a,b]\right)$ over intervals $[a,b]\in\left[\frac{r_\textrm{min}}{r_\textrm{max}},1\right)$.
We choose $r_\text{min}=-100$ and $r_\text{max}=10$.
While these bounds are arbitrary, they reflect the undesirability of failing in the task demonstrated by the expert.
The reward feature $\nu$ is a one-hot indicator over the states.
\Cref{fig:tt_scaling_constant} presents the agent's success rate in reaching the treasure, along with the average time exploring, which corresponds to the number of listening actions, for different values of the prior mean $k^\star$.
As expected from \Cref{sec:coe_prior}, when $k^\star$ approaches 1, the agent explores more, never exploiting when $k^\star=1$.
The probability of reaching the treasure increases with the number of listening actions.
Therefore, as $k^\star$ approaches 1 (without reaching it), the treasure rate also increases.
By contrast, when $k^\star$ decreases the agent begins to listen less often.
These behaviors correspond to the Bayes-optimal policies for each of the priors, demonstrating that our method learns the Bayes-optimal policy irrespective of the prior supplied.

\subsection{Investigating Latent Inference}
\label{subsec:investigating-latent-inference}
\begin{wrapfigure}{r}{0.5\textwidth}
\centering
\begin{minipage}{.245\textwidth}
\centering
\vspace{-4mm}
\includegraphics[width=\textwidth]{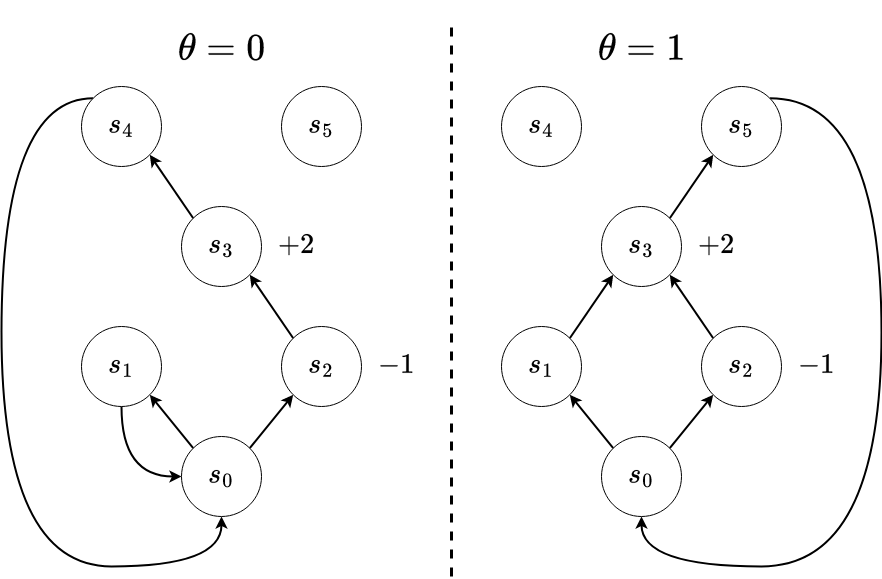}
\end{minipage}%
\begin{minipage}{.245\textwidth}
\centering
\includegraphics[width=\textwidth]{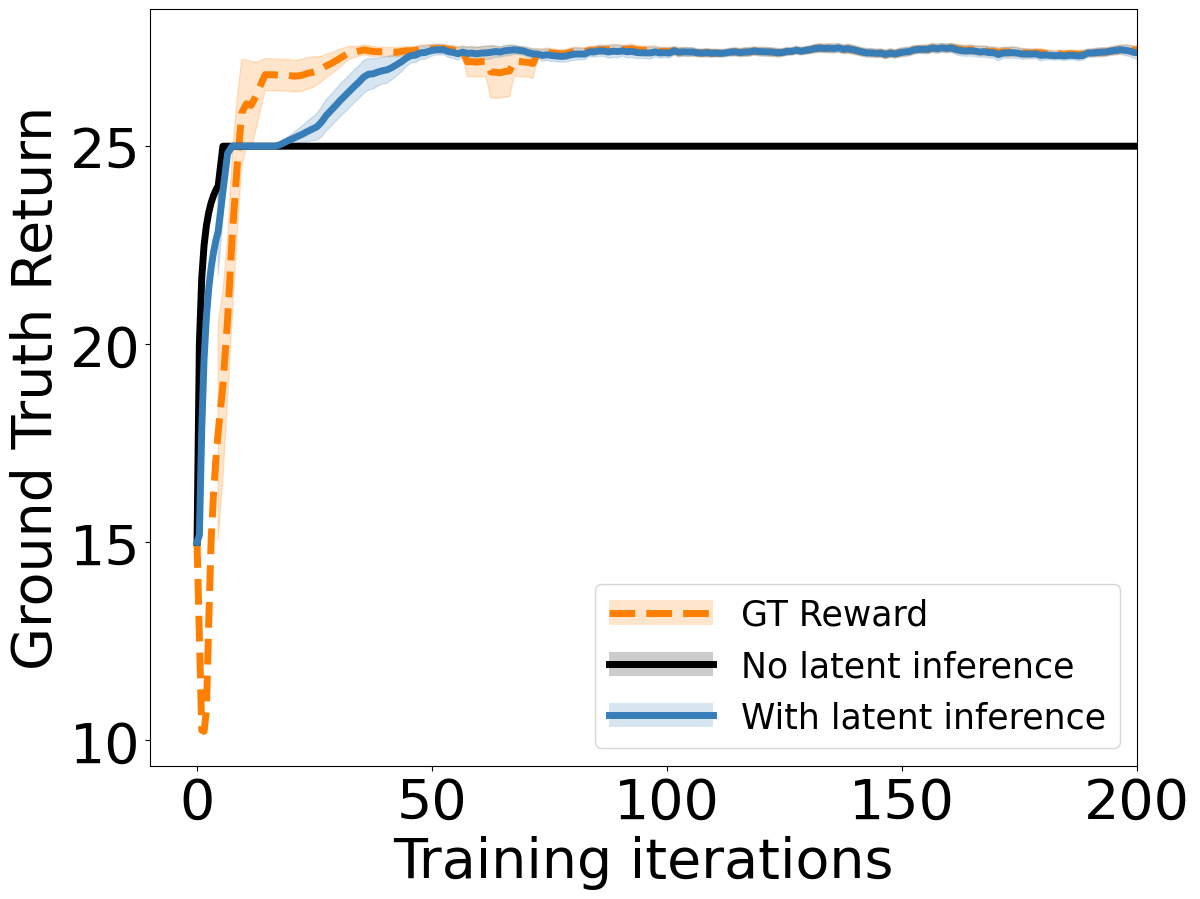}
\end{minipage}
\vspace{-2mm}
\caption{
\small
A demonstration of the necessity for latent inference with \ouralgo.
On the left, we show the CMDP used in the experiments, with two possibilities for the context $\theta$.
On the right, we show the ground truth returns of a DQN agent for trajectories of 100 steps in the CMDP during training.
The shading shows the standard error of the mean for 8 seeds.
}
\vspace{-2mm}
\label{fig:counterexample}
\end{wrapfigure}

Next, we look at whether inferring the latent $\theta$ during IRL matters for learning the desired reward function.
We experiment in a simple CMDP environment depicted in \Cref{fig:counterexample}.
For a mathematical example, see \Cref{app:further_counterexample}.
In this environment, the expert policy goes to the state $s_3$, which provides a reward of $+2$ and then loops back to state $s_0$ through $s_4$ or $s_5$.
It prefers to take the route through $s_1$, when it is available, to avoid the negative reward of $-1$ in $s_2$.
When $\theta$ is distributed such that $p(\theta = 0) > p(\theta = 1)$, it can lead to naive IRL misidentifying the expert intent.
To see why, consider that to fit the expert behavior without information about $\theta$, the reward function has to make the path through $s_2$ more likely in both MDPs.
Conditioning the successor features on the inferred $\theta$ resolves this issue, because then identifying the reward reduces to standard IRL in two separate MDPs.
To verify this in practice, we show results of learning the reward functions with and without latent inference in \Cref{fig:counterexample}.
We choose the latent to be distributed as $p(\theta = 0) = 0.9$.
No reward prior is used.
The policy trained on the rewards learned with latent inference matches the policy trained with the ground truth rewards.
The rewards learned without latent inference do not result in as good a policy, demonstrating that latent inference is necessary to learn the correct rewards in a general CMDP.
See \Cref{app:counterexample_experiment} for an analysis of the learned rewards.

\subsection{Reward Priors in a Larger CMDP}
\label{subsec:tiger-maze}
\begin{figure}
\centering
\includegraphics[width=0.3\textwidth]{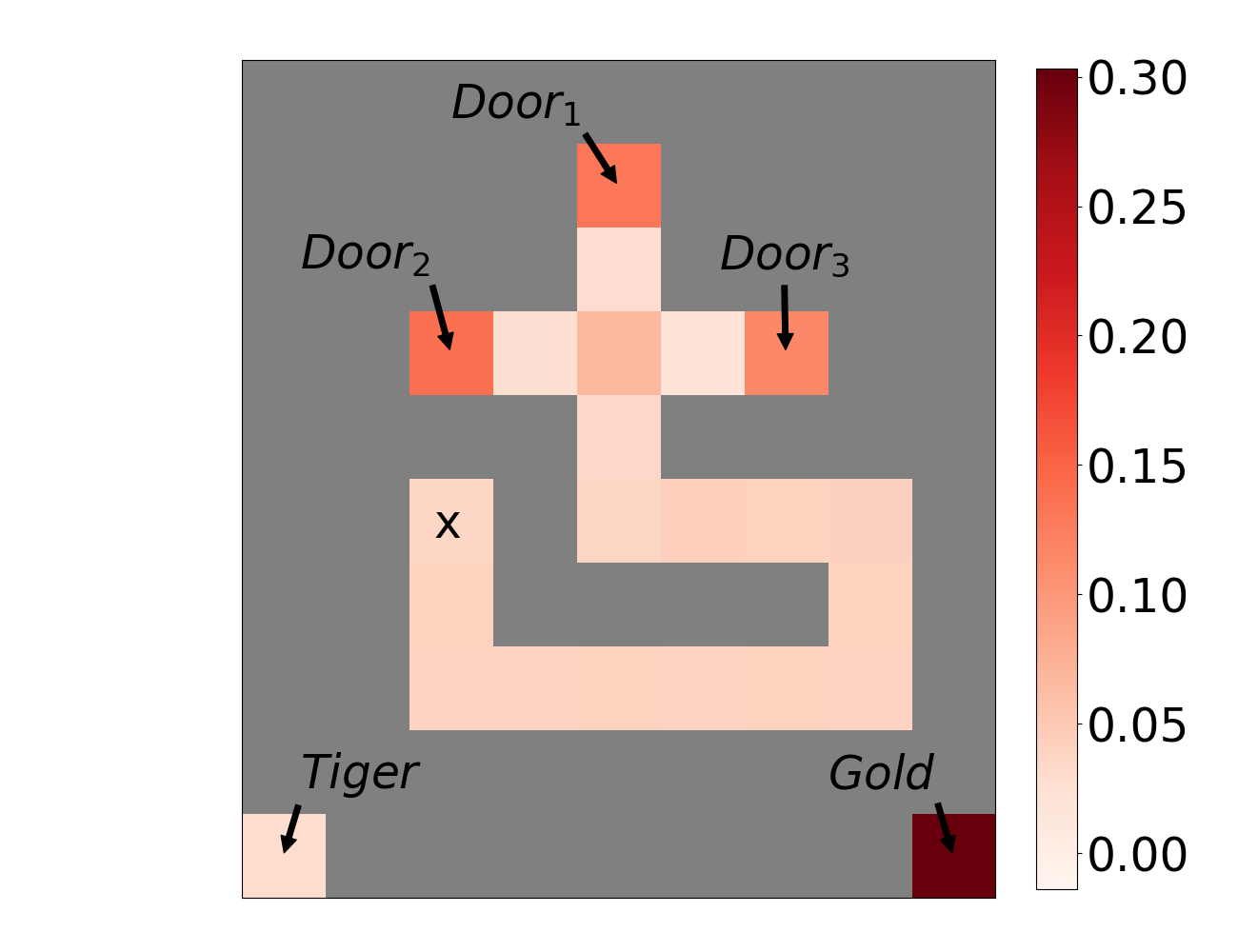}
\includegraphics[width=0.3\textwidth]{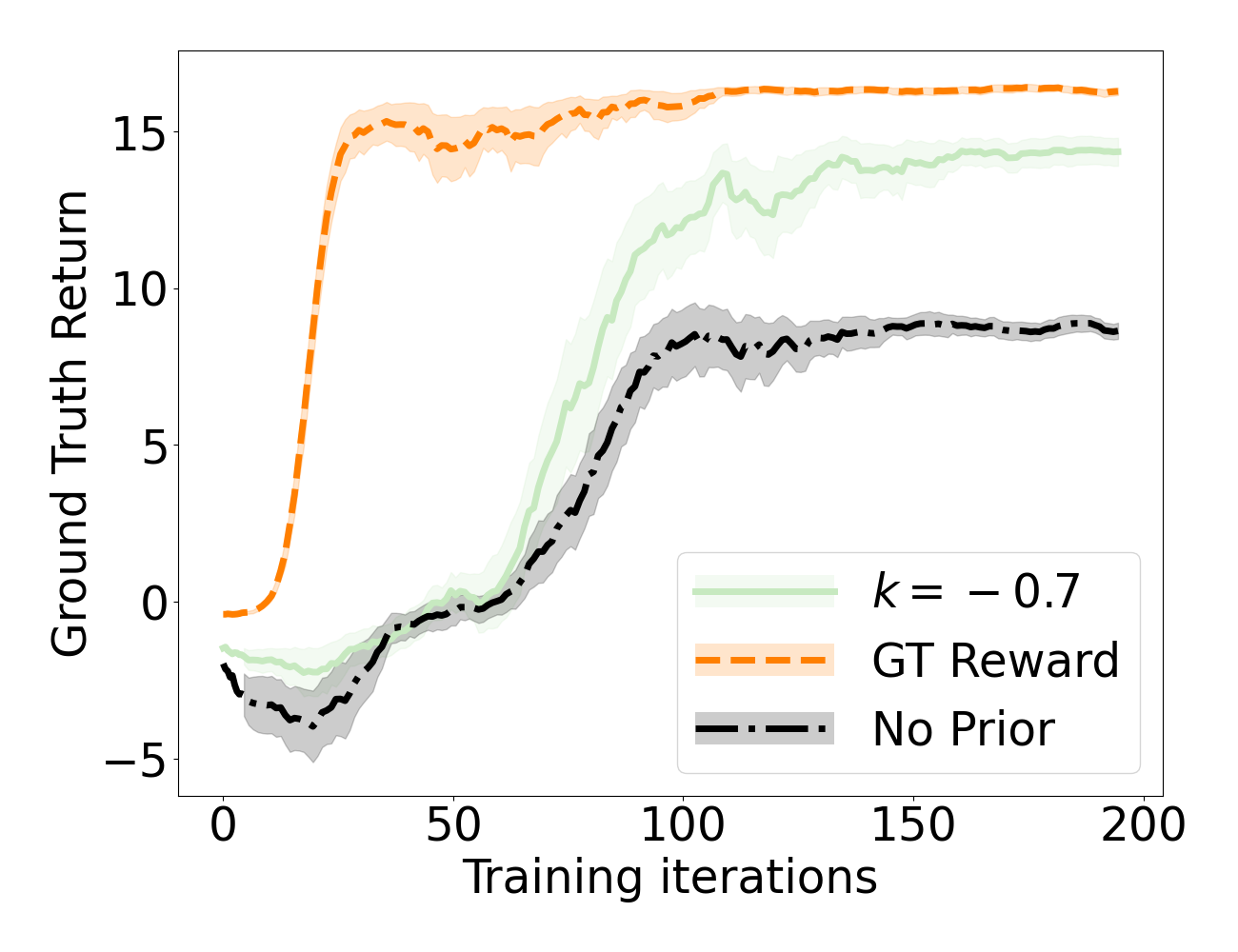}
\includegraphics[width=0.3\textwidth]{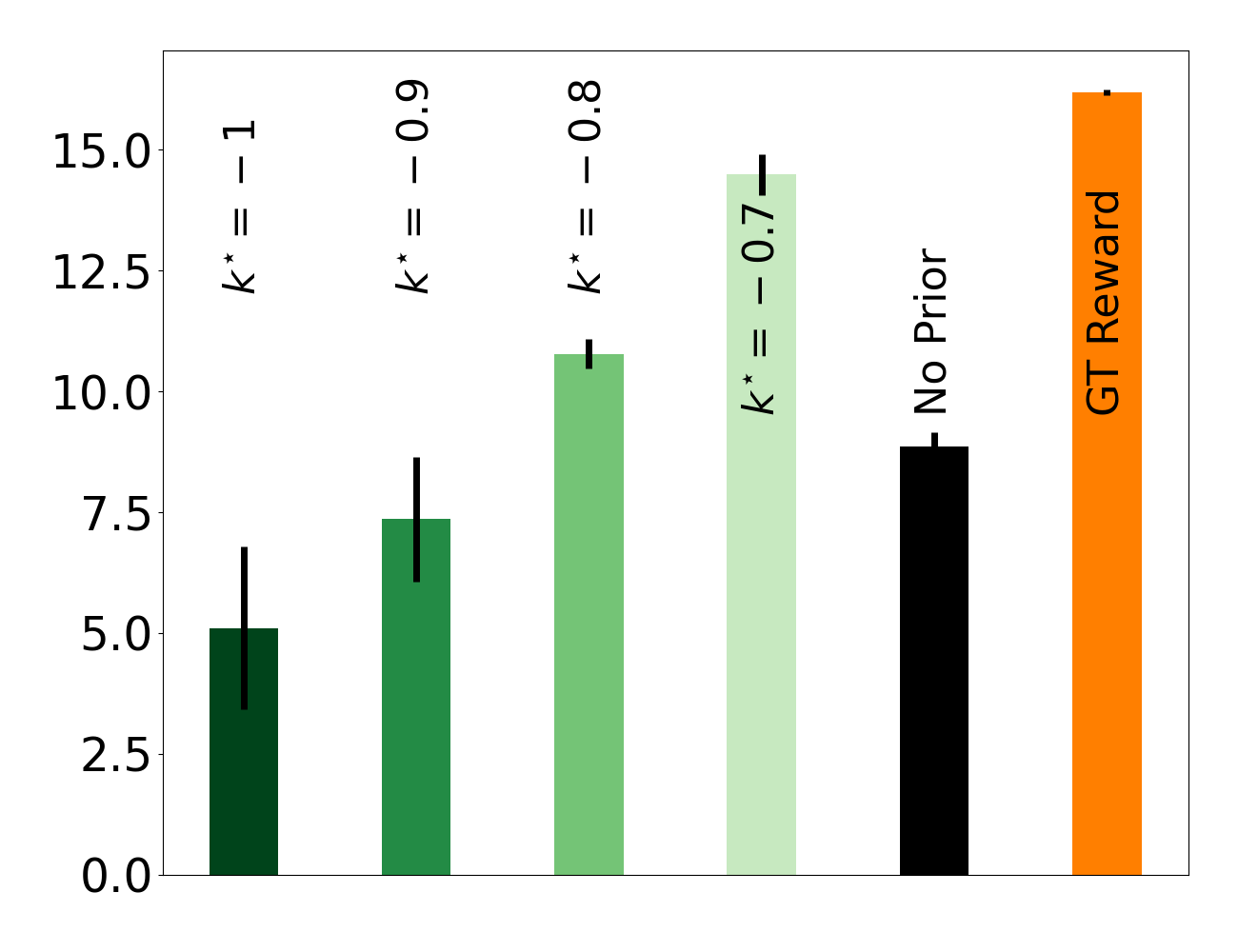}
\vspace{-2mm}
\caption{
\small
\ouralgo successfully learns the optimal behavior in a challenging gridworld environment.
On the left, we show the rewards learned by the contextual IRL.
In the middle, we show the return (using the manually constructed reward) of policies trained with reward inferred with and without a reward prior and the manually constructed reward (ground truth).
The shading shows the standard error of the mean for 8 random seeds.
On the right, we show the final returns of policies trained using different values of $k^\star$ compared to not using the reward prior and using the ground truth reward.
}
\vspace{-3mm}
\label{fig:tiger_maze}
\end{figure}

Finally, we test \ouralgo in a grid-world environment with pixel-based observations.
The agent observes a top-down view of the environment similar to the illustration of the learned rewards presented in \Cref{fig:tiger_maze}.
The agent can move in four directions and take listening actions.
Taking the listening action in any grid cell results in a stochastic transition to a state, which indicates the location of the gold, but otherwise has the same dynamics as the cell that the action was taken in.
When the agent enters a door, it is moved to $\mathrm{Tiger}$ or $\mathrm{Gold}$ depending on $\theta$.
From those states, the agent is moved to the grid cell marked with $x$ in the next timestep, setting the agent up for another round in the maze.
The expert takes the shortest path from any state to $\mathrm{Gold}$, choosing the correct door, depending on $\theta$.
See \Cref{app:tiger_maze} for details.

Three DQN training curves are shown in \Cref{fig:tiger_maze} corresponding to a manually constructed ground truth reward, which explains the expert behavior, reward learned by the contextual IRL without using a reward prior, and the same reward refined using the prior.
These learning curves show that \ouralgo with a particular reward prior produces a similar BAMDP policy as the manually constructed reward.
At the same time, using just the IRL reward results in a policy that initially chooses a door at random.
We compare multiple values for $k^\star$ and find that as in the simple Tiger-Treasure environment, different values result in over-exploration or under-exploration leaving a range in the middle that results in similar exploration as the handcrafted reward.
This experiment shows that \ouralgo can recover a reward function, which enables the agent to complete the task demonstrated by the expert despite a challenging imitation gap.

%% file: sections/related.tex
\section{Related Work}
\label{sec:related}

\textbf{Imitation Gaps.}
Prior work addressing the imitation gap typically assumes privileged access to the true environment reward during training.
In contrast, \ouralgo makes no such assumption.
\citet{weihs2021bridging, nguyen2022leveraging} assume access to the true reward during training, and propose to bridge the imitation gap by training on a weighted imitation and RL loss.
Other works propose to also integrate privileged information about the expert during training~\citep{chen2019lbc} in addition to the reward.
For example, Elf Distillation~\citep{walsman2023impossibly} studies an approach that mixes environment reward with online advice from the expert.
\citet{cai2021seeing} propose several stages of training, including those on privileged expert states, to connect the imitator and expert's observation spaces.
Separately, versions of the imitation gap have been considered by \citet{kwon2020inverse,Ortega2021-hh,swamy2022sequence,vuorio2022deconfounded} that assume that no new exploratory behavior needs to~be~learned.
The setting considered by \citet{straub2024probabilistic} is closer to our work but it only considers inferring unknown parameters from an agent acting under partial observability (that is, the IRL problem).
Our setting also requires an agent to act optimally in an \emph{unknown} environment \emph{without} a reward signal, i.e., there is uncertainty in the MDP at test time.

\textbf{Bayesian Reinforcement Learning.}
Our work shares similar components to other Bayesian approaches to RL.
For instance, VariBAD and related methods~\citep{zintgraf2019varibad, varibad_jmlr, hyperx} consider a model-based approach for learning approximations to Bayes-optimal policies by exploiting meta-reinforcement learning~\citep{beck2023survey} to perform inference over a subset of the unobserved context.
Similarly, BEN~\citep{Fellows23} is a model-free approach that learns Bayes-optimal policies by specifying a model and prior over the optimal Bellman operator.
All of these methods assume access to the reward and do not consider the issue of imitating an expert.
Bayesian Inverse RL approaches infer a posterior over the unknown reward distribution given trajectories of demonstrations~\citep{Ramachandran07} (see \citet{Adams22}, Table 1 for a complete list of existing approaches).
\ouralgo generalizes these approaches in \Cref{sec:inferring_reward} to account for the unobserved context variable before integrating the learned reward posterior into a BAMDP.

\textbf{Successor Features.}
Successor features \citep{Barreto17,Brown2019} provide an elegant representation of value functions under the assumptions of a linear reward function.
\citet{Janz19} incorporate successor features into the Bayesian framework.
Most similar to \ouralgo, successor features have been used successfully in Psi-Phi Learning~\citep{filos2021psiphilearning} for multi-task inverse reinforcement learning.
Psi-Phi Learning can also retrieve the reward parameterizations for a new agent (expert) but does so with full observability.

%% file: sections/conclusion.tex
\section{Limitations}
\label{sec:limitations}
To make the empirical setup close to the theory, we use ground truth inference models and linearity assumptions, which means that scaling up the algorithm requires revisiting those choices.
For example, by learning an approximate inference model.
In the CMDPs considered in this paper, we assumed the set of allowed exploration states was explicitly known a priori, which may be a limiting assumption in harder problems.
As is typical in Bayesian methods, we leave the design of problem-specific priors to the practitioners focused on those problems.

\section{Conclusion}
\label{sec:conclusion}
In this paper, we proposed a fully Bayesian solution to the imitation gap which integrated various priors over the reward parameterization, cost of exploration, and hidden state.
This allowed us to derive a BAMDP formulation of the problem whose solution optimally trades off exploration and exploitation at test time.
We demonstrated the importance of each component of our algorithm across a series of experiments with an imitation gap before showing that \ouralgo scales to larger maze problems, including those with high-dimensional pixel-based observations.
Crucially, in contrast to previous work, no online reward information was required.
This makes our work particularly exciting for challenging imitation problems where no reward is easily specifiable, and extending \ouralgo to even more complex settings is a promising direction for future work.

%% file: sections/appendix.tex
\section*{\LARGE \centering Supplementary Material}
\label{sec:appendix}

\section{Summary of Distributions in \ouralgo}
\label{appsec:dist_summary}
Due to the diverse nature of the sources of input information, inferring the reward posterior requires the agent to learn and maintain several distributions over random variables, we summarize them in \Cref{table:distributions}.

\begin{table*}[ht]
\vspace{-2mm}
\scriptsize
\caption{
\small
Summary of the distributions involved in \ouralgo. This table supplements the diagram in \Cref{fig:schematic}.}
\vspace{1mm}
	\label{table:distributions}
	\begin{center}
	\begin{tabular}{c c cc}
		\textbf{DISTRIBUTION} & \textbf{NAME} & \textbf{DESCRIPTION} & \textbf{SAMPLES TO LEARN}\\ \toprule 
$p(\omega)$ & \begin{tabular}{@{}c@{}} Reward \\ Parameter Prior \end{tabular}  &  \begin{tabular}{@{}c@{}} Incorporates prior knowledge in reward \\  parameterization \end{tabular} & None \\  \\
	$p(k)$ &  \begin{tabular}{@{}c@{}} Cost of\\ Exploration Prior \end{tabular}  & \begin{tabular}{@{}c@{}}Incorporates prior knowledge in rewards  \\$r(s,a)=k r_\textrm{max}$ over exploratory state-actions \end{tabular} & None  \\ \\
$p(\theta)$ & Contextual Prior &  \begin{tabular}{@{}c@{}}Characterizes uncertainty in $\theta$ a priori at\\ test time\end{tabular} & \begin{tabular}{@{}c@{}}$\mathcal{D}_\textrm{Sim}$ - samples from  the CMDP \\simulator \end{tabular}  \\ \\
$p(a\vert s,\theta ,\omega )$ &   \begin{tabular}{@{}c@{}} Model \\ Expert Policy \end{tabular} &  \begin{tabular}{@{}c@{}} Model of optimal policy for expert  in\\ CMDP $\mathcal{M}(\theta)$  with reward parameters $\omega$ \end{tabular} & \begin{tabular}{@{}c@{}} Contextual successor features\\ learned from simulator and $\mathcal{D}_\textrm{Expert}$ \end{tabular} \\  \\
					$p(\omega\vert \mathcal{D}_\textrm{Expert})$ &  \begin{tabular}{@{}c@{}} Expert reward \\ Posterior \end{tabular}  & \begin{tabular}{@{}c@{}}Updates the reward parameter prior using \\ expert data \end{tabular} & \begin{tabular}{@{}c@{}}$ \mathcal{D}_\textrm{Expert}$   dataset of expert\\ trajectories \end{tabular}   \\ \\
		$p(\theta_i\vert \tau_i)$ & \begin{tabular}{@{}c@{}}Contextual\\ Posterior\end{tabular} &  \begin{tabular}{@{}c@{}} Characterizes uncertainty    over which\\ latent variable  $\theta_i$  agent $i$ was assigned \end{tabular} &  \begin{tabular}{@{}c@{}} $\tau_i$ - each expert or exploratory \\ agent's trajectory  \end{tabular}  \\  \\
				$p^\textrm{IRL+COE}_\textrm{Bayes}(r\vert s,a)$ & \begin{tabular}{@{}c@{}}Bayesian Reward\\ Distribution\end{tabular} &  \begin{tabular}{@{}c@{}} Incorporates epistemic uncertainty from \\ $p(\omega\vert \mathcal{D}_\textrm{Expert})$ and $p(k)$ into reward model \end{tabular} &  \begin{tabular}{@{}c@{}} None  \end{tabular}  \\ 
		\bottomrule
\end{tabular}
\end{center}
\vspace{-2mm}
\end{table*}

\section{Bayesian Successor Feature Learning}
\label{app:bsfl}

Consider the optimal $Q$-function $Q^\star(s,a,\theta,\omega)$, which satisfies the optimal contextual Bellman equation: $\mathcal{B}^\star[Q^\star](s,a,\theta,\omega)=Q^\star(s,a,\theta,\omega)$ where: %$\mathcal{B}^\star[Q^\star](s,a,\theta,\omega)$ is the optimal contextual Bellman operator:
\begin{align}
	&\mathcal{B}^\star[Q^\star](s,a,\theta,\omega)\coloneqq \nu(s,a)^\top\omega\\
	&\qquad+\gamma \mathbb{E}_{s'\sim p(s'\vert s,a,\theta)}\left[\sup_{a'} Q^\star(s',a',\theta,\omega)\right].
\end{align}
As the expert selects actions $a\in \argmax_{a'}Q^\star(s,a',\theta,\omega)$, learning $Q^\star(s,a,\theta,\omega)$ is sufficient for modeling the set of expert policies, from which the true reward parametrization can be inferred. 

\textbf{Learning with Expert Data.}
Consider the Bellman equation under the expert policy $\pi^\star(a'\vert s',\theta_i)$.
The successor feature representation $\Psi_\phi$ should satisfy:
\begin{align}
	\Psi_\phi(s,a,\theta_i)^\top \omega^\star=\nu(s,a)^\top\omega^\star+\gamma\mathbb{E}_{s'\sim p(s'\vert s,a,\theta_i),a'\sim \pi^\star(a'\vert s',\theta_i) }\left[\Psi_\phi(s',a',\theta_i)\right]^\top \omega^\star.
\end{align}
We can factor $\omega^\star$ from the Bellman equation and then solve:
\begin{align}
	\Psi_\phi(s,a,\theta_i)=\nu(s,a)+\gamma\mathbb{E}_{s'\sim p(s'\vert s,a,\theta_i),a'\sim \pi^\star(a'\vert s',\theta_i) }\left[\Psi_\phi(s',a',\theta_i)\right].
\end{align}
This yields the objective for each expert trajectory $\tau_i$: 
\begin{align}
	\mathcal{L}_\textrm{Expert}(\phi;\tau_i)\coloneqq \mathbb{E}_{s,a\sim\textrm{Unif}(\tau_i), \theta_i\sim p(\theta_i\vert \tau_i)} \big[\lVert &\Psi_\phi(s,a,\theta_i)-(\nu(s,a) \\ &+\gamma\mathbb{E}_{s'\sim p(s'\vert s,a,\theta_i),a'\sim \pi^\star(a'\vert s',\theta_i) }\left[\Psi_\phi(s',a',\theta_i)\right]) \rVert\big]~\label{eq:sf_expert_objective}.
\end{align}
where $\textrm{Unif}(\tau_i)$ is a uniform distribution over the state-action pairs in trajectory $\tau_i$.
Minimizing $\mathcal{L}_\textrm{Expert}(\phi;\tau_i)$ for each trajectory ensures that the successor feature representation is consistent with the expert demonstrations.
Minimizing $\mathcal{L}_\textrm{Expert}(\phi;\tau_i)$ can be carried out using a semi-gradient approach, thereby avoiding the need for two samples from the expert policy and state-transition distribution:
\begin{align}
	&\phi\leftarrow \phi+\eta_\phi \nabla_\phi \Psi_\phi(s,a,\theta_i) (\nu(s,a)+\gamma\Psi_\phi(s',a',\theta_i)-\Psi_\phi(s,a,\theta_i)).
\end{align}
As the latent contextual variable $\theta_i$ is never observed, we must infer a posterior over its value $p(\theta_i\vert\tau_i)$. It may seem that this could be avoided via IRL using the prior-averaged transitions $\mathbb{E}_{\theta\sim p(\theta)}\left[p(s'\vert s,a,\phi)\right]$. However, we provide a simple counterexample in \Cref{app:further_counterexample} demonstrating that a prior-averaged approach does not account for the true reward ordering in the underlying space of CMDPs.
Sampling $\theta_i\sim p(\theta_i\vert\tau_i)$ from the posterior over the expert's contextual variable is typically intractable, so we use variational inference instead, as detailed in \Cref{app:approximate_inference}.

\textbf{Learning with a Simulator.}
In addition to expert demonstrations, we also have access to samples from the CMDP simulator.
This allows us to learn $\Psi_\phi$ over state-action pairs where the expert has not provided demonstrations.
We sample a dataset $\mathcal{D}_\textrm{Simulator}\coloneqq \{\tau_i\}_{i=1}^{N_\textrm{Simulator}}$ of $N_\textrm{Simulator}$ trajectories from the simulator: the simulator samples an MDP from the prior, and then an exploration policy $\pi_\textrm{Explore}$ interacts with the corresponding MDP, observing state-action-state transitions.
In this paper, we use an $\epsilon$-greedy exploration policy where the greedy actions $a'\in\argmax_{a'} \Psi_\phi(s', a', \theta_i)^\top \omega$ are taken with probability $1-\epsilon$ then uniformly otherwise.
More sophisticated approaches such as a policy that maximizes the entropy of the ergodic, discounted state-action occupancy distribution~\citep{Hazan19} would also be appropriate, especially in larger domains. However, our experiments demonstrate the $\epsilon$-greedy policy suffices for learning successor features in our setting.

Once samples have been obtained, we infer the posterior $p(\theta_i\vert \tau_i)$ over the MDP that the exploratory agent was assigned for each MDP $i\in[1:N_\textrm{Simulator}]$.
Like with the expert data, our goal is to ensure that the successor feature representation $\Psi_\phi$ satisfies a Bellman equation.
In the target, we choose the next action that maximizes the $Q$-function for a given $\omega$, $a'\in\argmax_{a'} \Psi_\phi(s', a', \theta_i)^\top \omega$.
This yields the objective for each MDP $i$:
\begin{align}
	\mathcal{L}_\textrm{Simulator}(\phi;\tau_i,\omega)\coloneqq~& \mathbb{E}_{s,a\sim\textrm{Unif}(\tau_i), \theta_i\sim p(\theta_i\vert \tau_i)} \big[\\ &\big(\Psi_\phi(s,a,\theta_i) -(\nu(s,a) + \gamma \mathbb{E}_{s'\sim p(s'\vert s,a,\theta_i) }\left[\Psi_\phi(s',a',\theta_i)\right] \big)^2\big] \label{eq:bellman_eq_for_sf}.
\end{align}
where $\textrm{Unif}(\tau_i)$ is a uniform distribution over the state-action pairs in trajectory $\tau_i$.
We optimize this objective using the following TD update:
\begin{align}
	\phi\leftarrow \phi+\eta_\phi \nabla_\phi \Psi_\phi(s,a,\theta_i)\cdot (\nu(s,a)+\gamma\Psi_{\phi'}(s',a',\theta_i)-\Psi_\phi(s,a,\theta_i)).
\end{align}
There are two differences between the updates for the exploratory data and the expert data. First, the exploratory data updates are off-policy.
Due to the deadly triad \citep{Sutton1998}, using semi-gradients is not guaranteed to converge \citep{Tsitsiklis97,Fellows23}.
We introduce a separate target network $\phi'$ that is updated periodically to stabilize the updates. Second, the updates depend on $\omega$ as the supremum acts over the dot product between the successor representation and the reward parametrization.
As we require successor features to infer $\omega$, we interleave learning both $\omega$ and $\phi$ in a nested optimization, using both expert and exploratory data with an initial burn-in period using only the expert data.

We combine both objectives into the single objective presented in \cref{sec:contextual_successor_features}:
\begin{align}
	\mathcal{L}(\phi;\omega)=\mathcal{L}_\textrm{Expert}(\phi;\tau_i)+\beta\mathcal{L}_\textrm{Simulator}(\phi;\tau_i,\omega),
\end{align}
for some constant $\beta$. 
However, we note that the reward learning and successor feature learning depend on each other through the objective $\mathcal{L}_\textrm{Simulator}$ and reward update given by \Cref{proof:laplace_parameter}, yielding a two-timescale learning problem.
To enable convergence to a stable local optimum, the learning rates $\eta_{\phi}$ and $\eta_{\omega}$ are set such that $\eta_{\omega} < \eta_{\phi}$.

\section{The Need for Inference Over Context Variables}
\label{app:further_counterexample}

\begin{figure}
\begin{center}
    \includegraphics[width=0.5\textwidth]{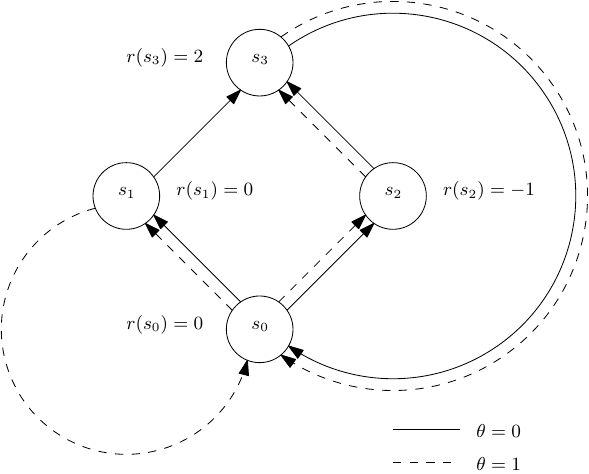}
\end{center}
\caption{Counterexample CMDP}
\label{fig:math_counterexample}
\vspace{-4mm}
\end{figure}
We consider a simple counterexample illustrated in \Cref{fig:math_counterexample} with four states $s_i$ and two possible hidden contexts denoted by $\theta$.
As shown in the figure, $\theta$ controls the environment transition.
The simplest approach to IRL using the average MDP would map the experts' state visitation frequency $\eta$ to reward value.
This would assign rewards: $r(s_1)=c \eta  $, $r(s_2)=c( 1-\eta)  $, $r(s_3)=c $, where $c$ is an arbitrary finite positive constant.
This means that the values of $r(s_1)$ and $r(s_2)$ are only determined by $\eta$ and will give incorrect reward ordering  $r(s_2)\ge  r(s_1)$ for any $\eta\le 0.5$.
On the other hand, if a Bayesian approach is taken, it is possible to infer which MDP the agent was in. If the belief of a trajectory is weighted towards the expert being in $\theta=0$, the expert's preference for $a_0= \textrm{left}$ over $a_1= \textrm{right}$ must be a consequence of the reward ordering $r(s_1)>r(s_2)$.
Likewise, if the belief of a trajectory is weighted towards the expert being in $\theta=1$, the expert's preference for $a_0= \textrm{right}$ must imply that $r(s_2)+\gamma r(s_3)> r(s_1)+\gamma r(s_0)$. As $r(s_1)>r(s_2)$ will be inferred from the trajectories of experts in $\theta=0$, this implies that $r(s_0)<r(s_3)$, and so a correct reward ordering will be learned regardless of $\eta$.
By being Bayesian, these possible inferences condition on which $\theta_i$ the expert was in, and are explained by $Q_{\zeta^\star}(s,a,\theta_i,\omega)$.
We can best match   $\omega$ so that is consistent with these inferences under the belief $p(\theta_i\vert \tau_i, \omega)$ in the expert's MDP.
We thus conclude that compared to a fully Bayesian approach, naively using imitation learning on the prior-averaged MDP will not yield policies that account for reward ordering of the underlying MDP.

\section{Bayesian Solution to the Imitation Gap}
\label{app:main_algorithm}
Finally, we present pseudocode for the Bayesian solution to the Imitation Gap (\ouralgo) in \Cref{alg:big}.
It takes as inputs the prior distributions and the expert dataset and produces a reward function $r_\textrm{Bayes}^\text{IRL + COE}$.
In the main loop, data is collected from the simulated environment using an $\epsilon$-greedy policy.
We use $\epsilon$-greedy for convenience as we found it to be an easy way to achieve suitable coverage of the state-action space in the experiments.
The data collected using the random policy is used for learning the successor features in an off-policy RL algorithm.
With an off-policy algorithm, assuming coverage of the whole state-action space, the exact kind of randomness does not matter for the kind of successor features we learn.
The collected data together with the expert demonstrations are used to update the successor feature representation.
The learned successor features are used for updating the reward parameters to maximize the likelihood of the expert data.
After $K$ steps of the main loop, the final reward function is computed by applying the cost-of-exploration refinement defined in \Cref{sec:coe_prior}.

\begin{figure}[t!]
\begin{algorithm}[H]
\small
\caption{\textsc{\small Bayesian Solution to the Imitation Gap}}
\label{alg:big}
\begin{algorithmic}
\REQUIRE priors $p(\theta)$, $p(\omega)$, and $p_\textrm{COE}(r\vert \sigma,s,a)$, contextual posterior $p(\theta \vert \tau)$, dataset $\mathcal{D}_\textrm{Expert}$, reward scales $r_\textrm{min}$ and $r_\textrm{max}$, learning rates $\eta_{\omega}$ and $\eta_{\phi}$, loss coefficient $\beta$, and number of steps $K$.
\STATE Initialize parameters $\phi$ for the successor features and $\omega$ for the reward.
\STATE Initialize an empty replay buffer $\mathcal{D}_\textrm{Replay}$.
\FOR {$K$ steps}
\STATE Sample a new MDP from simulator $\theta_i \sim \rho(\theta)$.
\STATE Sample a trajectory $\tau_i$ in the MDP defined by $\theta_i$ using an epsilon-greedy policy w.r.t.\ the $Q$-function defined by $\Psi_{\phi}(s, a, \tilde \theta_i)^\top \omega$ for $\tilde \theta_i \sim p(\theta_i | \tau_i)$. 
\STATE Add the trajectory $\tau_i$ to the replay buffer $\mathcal{D}_\textrm{Replay}$.
\STATE Update $\phi$ to minimise $\mathcal{L}(\phi;\omega)$ using $\mathcal{D}_\textrm{Expert}$ and $\mathcal{D}_\textrm{Replay}$.
\STATE Update $\omega$ using  \Cref{eq:stochastic_updates} with $\mathcal{D}_\textrm{Expert}$.
\ENDFOR
\STATE Rescale $\omega$ s.t. rewards lie in the range $[r_\textrm{min},r_\textrm{max}]$.
\STATE Compute the predictive reward  $r_\textrm{Bayes}^\text{IRL + COE}$ .
\RETURN $r_\textrm{Bayes}^\text{IRL + COE}$
\end{algorithmic}
\end{algorithm}
\end{figure}

\section{Approximate Inference}
\label{app:approximate_inference}

Whilst the full Bayesian approach outlined in \Cref{sec:full_Bayes} is clearly desirable, there are several sources of intractability that prevent us from computing the Bayes-optimal policy $\pi^\star_\textrm{Bayes}$ exactly. Firstly, maintaining and inferring the posterior distributions is likely to be intractable for all but the simplest choice of likelihoods, which have insufficient capacity for representing the set of MDPs beyond contrived toy tasks. Secondly, marginalization using the posteriors likely will involve high dimensional integrals, which are computationally inefficient. Finally, solving the planning problem in the BAMDP for every possible history to obtain a Bayes-optimal policy is notoriously challenging, even for very simple domains \citep{zintgraf2019varibad, varibad_jmlr}. We thus derive an algorithm that follows the methodology of the formal Bayesian approach outlined in \Cref{sec:full_Bayes} whilst making necessary approximations from the powerful toolkit of variational inference to ensure tractability.

\subsection{Tractable Prior and Likelihood Learning}
\label{sec:VI_for_prior_learning}
Instead of attempting to infer the posterior $p(\phi\vert \mathcal{D}_\textrm{simulator})$ exactly, which may be intractable, we use a MAP approach instead to learn a point estimate $\phi^\star_\textrm{MAP}$. The justification for this is that we have access to a simulator from which a large number of samples can be drawn. The Bernstein-von Mises theorem specifies that in the limit of large data $K\rightarrow\infty$, $p(\phi\vert \mathcal{D}_\textrm{simulator})\rightarrow \delta_{\phi^\star_\textrm{MLE}}(\phi)$ where $\phi^\star_\textrm{MLE}$ is the maximum likelihood estimator, so we expect $p(\phi\vert \mathcal{D}_\textrm{simulator})\approx \delta_{\phi^\star_\textrm{MLE}}(\phi)$. In the same limit, $\phi^\star_\textrm{MAP}\rightarrow \phi^\star_\textrm{MLE}$. Our choice of using a MAP estimator over the MLE estimator is purely practical in that the prior can add regularization to stabilize learning, as is seen by the contribution to the MAP objective:
\begin{align}
	\mathcal{L}_\textrm{MAP}(\phi)=\sum_{i=1}^K\log p(\tau_i\vert \phi) +\log(\phi).\label{eq:map_prior}
\end{align}
To find a tractable objective to maximize the MAP estimate, we introduce a variational distribution $q_{\chi_i}(\theta_i)$ parametrized by ${\chi_i}\in X$, for which:
\begin{align}
	\log p(\tau_i\vert \phi)&= \int \log  p(\tau_i\vert \phi) q_{{\chi_i}} (\theta_i)d\theta_i,\\
	&= \int \log \left( \frac{p(\tau_i,\theta_i\vert \phi)}{p(\theta_i\vert \tau_i,\phi)} \right)q_{{\chi_i}} (\theta_i)d\theta_i,\\
	&= \int \log \left( \frac{p(\tau_i,\theta_i\vert \phi)}{p(\theta_i\vert \tau_i,\phi)} \cdot \frac{q_{{\chi_i}} (\theta_i)}{q_{{\chi_i}} (\theta_i)}\right)q_{{\chi_i}} (\theta_i)d\theta_i,\\
	&= \int \left(\log p(\tau_i,\theta_i\vert \phi)-\log q_{{\chi_i}} (\theta_i) \right) q_{{\chi_i}} (\theta_i)d\theta_i-\int\frac{ \log p(\theta_i\vert \tau_i,\phi)}{\log q_{\chi_i} (\theta_i)}q_{\chi_i} (\theta_i)d\theta_i,\\
	&= \mathcal{L}_\textrm{ELBO}(\phi,{\chi_i})+\kl{q_{\chi_i}}{p(\cdot\vert \tau_i,\phi)},\\
	\implies \mathcal{L}_\textrm{ELBO}(\phi,{\chi_i})&=\log p(\tau_i\vert \phi)-\kl{q_{\chi_i}}{p(\cdot\vert \tau_i,\phi)},\label{eq:elbo_decomposition}
\end{align}
where $\mathcal{L}_\textrm{ELBO}(\cdot)$ denotes an evidence lower bound (ELBO). Now, assuming there exists some ${\chi_i}^\star\in X$ such that $q_{{\chi_i}^\star}(\theta_i)=p(\theta_i\vert \tau_i,\phi)$, then $\sup_{{\chi_i}\in X} \kl{q_{\chi_i}}{p(\cdot\vert \tau_i,\phi)}=0 $, hence it follows from \Cref{eq:elbo_decomposition} that
\begin{align}
	\log p(\tau_i\vert \phi)=\sup_{{\chi_i}\in X} \mathcal{L}_\textrm{ELBO}(\phi,{\chi_i}).
\end{align}
Substituting into the MAP objective from \Cref{eq:map_prior} yields:
\begin{align}
	\mathcal{L}_\textrm{MAP}(\phi)=\sum_{i=1}^K\sup_{{\chi_i}\in X} \mathcal{L}_\textrm{ELBO}(\phi,{\chi_i}) +\log(\phi).
\end{align}

To find a $\phi^\star_\textrm{MAP}\in \argsup_{\phi\in\Phi}\mathcal{L}_\textrm{MAP}(\phi)$, we recognize that for each MDP sampled from the simulator indexed by $i\in [1:K]$, we can minimize the following index-specific ELBO:
\begin{align}
	\mathcal{L}_\textrm{ELBO}(\phi,{\chi_i})=\mathbb{E}_{\theta_i\sim q_{\chi_i}(\theta_i)}\left[\sum_{t=1}^{H_i}\log p(s_t\vert s_{t-1},a_{t-1},\theta_i)+\log p(\theta_i\vert \phi) - \log(q_{\chi_i}(\theta_i))\right],
\end{align}
with respect to $\phi$ and $\chi_i$. Here, Variational Auto-encoders \citep{kingma2013auto} (VAEs) offer a powerful framework for solving this problem via a variational expectation-maximization (EM) algorithm. In VAE parlance, the distribution $p(\tau_i\vert \theta_i,\phi )$ is known as the \emph{decoder}, parametrized by $\phi\in\Phi$. 
% \begin{wrapfigure}{l}{0.52\textwidth}
	% \centering
	% \begin{minipage}{0.52\textwidth}
	% 	\vspace{-0.7cm}
		\begin{algorithm}[H]
			\caption{\textsc{LearnPrior+Likelihood}}
			\label{alg:LearnPrior}
			\begin{algorithmic}
				\STATE Initialize $\zeta,\chi,\phi $
				\FOR {$K$ steps}
				\STATE Sample new MDP from simulator $\theta\sim \rho(\theta)$
				\STATE Sample initial state $s_0 \sim p_0(s_0)$
				\FOR{$t\in[1:H_i]$}
				\STATE Sample action $a_{t-1}\sim d(a_{t-1})$
				\STATE Sample transition $s_t\sim p(s_t\vert s_{t-1},a_{t-1},\theta)$
				\STATE Sample gradient $g_{\chi} \sim\nabla_\psi \mathcal{L}_\textrm{ELBO}(\phi,\chi)$
				\STATE $\chi \leftarrow \chi + \alpha_\chi^t g_{\chi}$
				\STATE Sample gradient
				\STATE $g_{\phi} \sim \nabla_\phi\left(\mathcal{L}_\textrm{ELBO}(\phi,\chi)+\frac{1}{K H_i} \log p(\phi)\right)$
				\STATE $\phi \leftarrow \phi + \alpha_\phi^t g_\phi$
				\STATE $g_{\zeta} \sim \nabla_\zeta\textrm{MSBE}(\zeta)$
				\STATE $\zeta \leftarrow \zeta - \alpha_\zeta^t g_\zeta$
				\ENDFOR
				\ENDFOR
				\RETURN $\zeta,\phi$
			\end{algorithmic}
		\end{algorithm}
	% 	\vspace{-0.9cm}
	% \end{minipage}
% \end{wrapfigure}

For each MDP $\theta_i$, we train an \emph{encoder} $q_{\chi_i}(\theta_i)$, parametrized by $\chi_i\in X$, that acts as a variational approximation to the posterior: $p(\theta_i\vert \tau_i,\phi)$.

To learn the prior parameters, we propose \Cref{alg:LearnPrior}.
Instead of specifying an encoder for each $q_{\chi_i}(\theta_i)$, we train a single encoder $q_{\chi}(\theta_i)$ online using an entire trajectory of samples from a specific MDP $\theta_i$. Once training has finished for that MDP, we sample another $\theta_{i+1}$, using $\chi$ as the initialization parameters for the new encoder, $q_{\chi}(\theta_{i+1})$.
Moreover, as both learning the reward prior and minimizing the MSBE to learn the likelihood require samples from a simulator, we interleave both optimization problems using the same interactions with the simulator.

\section{Proofs and Derivations}
\label{app:proofs}
\renewcommand{\thesection}{\arabic{section}}
\setcounter{section}{4}
\setcounter{theorem}{0}

\begin{lemma}
	\label{proof:normalisation_gradient}
	The gradient of the log normalization constant $\log z(s,\omega,\theta_i)$ is 
	\begin{align}
		\nabla_\omega \log z(s,\omega,\theta_i)=\mathbb{E}_{a\sim p(a\vert s,\omega,\theta_i)}\left[ \Psi^\star(s,a,\theta_i) \right].
	\end{align}
	\begin{proof}
		We assume that $\mathcal{A}$ is continuous. If $\mathcal{A}$ is discrete, we replace the Lebesgue measure $\lambda$ with the counting measure, and our proof remains unchanged. Taking derivatives directly yields our desired result:
		\begin{align}
			\nabla_\omega \log z(s,\omega,\theta_i)&=\nabla_\omega 
			\log \int_\mathcal{A} \exp\left(\frac{1}{\alpha}\Psi^\star(s,a,\theta_i)^\top \omega\right) d\lambda(a),\\
			&=\nabla_\omega \int_\mathcal{A} \exp\left(\frac{1}{\alpha}\Psi^\star(s,a,\theta_i)^\top \omega\right) d\lambda(a)\cdot \frac{1}{ \int_\mathcal{A} \exp\left(\frac{1}{\alpha}\Psi^\star(s,a,\theta_i)^\top \omega\right) d\lambda(a)},\\
			&= \int_\mathcal{A} \Psi^\star(s,a,\theta_i) \frac{\exp\left(\frac{1}{\alpha}\Psi^\star(s,a,\theta_i)^\top \omega\right) }{ \int_\mathcal{A} \exp\left(\frac{1}{\alpha}\Psi^\star(s,a,\theta_i)^\top \omega\right) d\lambda(a)}d\lambda(a),\\
			&= \int_\mathcal{A} \Psi^\star(s,a,\theta_i) p(a\vert s,\omega,\theta_i)d\lambda(a),\\
			&=\mathbb{E}_{a\sim p(a\vert s,\omega,\theta_i)}\left[ \Psi^\star(s,a,\theta_i) \right].
		\end{align}
	\end{proof}
\end{lemma}
\setcounter{theorem}{0}
\begin{theorem} Define $\varsigma_0^2 \coloneqq \frac{\sigma_0^2}{\alpha}$. Using the Laplace approximation for the posterior, the approximate Bayesian reward parametrization $\omega^\star_\textrm{Bayes}\approx\omega^\star_\textrm{Laplace}$ can be found  by carrying out the following stochastic gradient descent updates on the log-posterior:
	\begin{align}
		\omega \leftarrow &\omega + \eta_\omega \left( N_\textrm{Expert}{H_i} \bigg(\Psi^\star(s_j,a_j,\theta_i)-\mathbb{E}_{a_i\sim p(a_i\vert s_j,\omega,\theta_i)}\left[ \Psi^\star(s_j,a_i,\theta_i) \right]\right)-\frac{(\omega-\omega_0)}{\varsigma_0^2}\bigg).
\end{align}
\begin{proof}
Using Laplace's approximation, we fit a Gaussian to the posterior distribution with mean $\omega^\star_\textrm{Laplace} \in \argsup_{\omega\in\Omega} p(\omega\vert \mathcal{D}_\textrm{Expert})$. Equivalently, we can maximize the log posterior instead. Defining $g(\omega;\mathcal{D}_\textrm{Expert})\coloneqq \nabla_\omega \log p(\omega\vert \mathcal{D}_\textrm{Expert})$, from the definition of the gradient of a log:
\begin{align}
	g(\omega;\mathcal{D}_\textrm{Expert})=\nabla_\omega \log p(\omega \vert \mathcal{D}_\textrm{Expert})=\nabla_\omega p(\omega \vert \mathcal{D}_\textrm{Expert})\cdot\frac{1}{p(\omega \vert \mathcal{D}_\textrm{Expert})}.\label{eq:log_gradient}
\end{align}
We define the joint set of expert contextual variables as $\Theta_\textrm{Expert}\coloneqq\{\theta_1,\theta_2,...\theta_{N_\textrm{Expert}}\}\in\Theta^{N_\textrm{Expert}}$. Taking gradients of the posterior yields:
\begin{align}
		\nabla_\omega p(\omega\vert \mathcal{D}_\textrm{Expert})&=\int_{\Theta^{N_\textrm{Expert}}} \nabla_\omega p(\omega\vert \mathcal{D}_\textrm{Expert},\Theta_\textrm{Expert})dP(\Theta_\textrm{Expert}\vert \mathcal{D}_\textrm{Expert}),\\
		&=\int_{\Theta^{N_\textrm{Expert}}}\frac{\nabla_\omega p(\omega\vert \mathcal{D}_\textrm{Expert},\Theta_\textrm{Expert})}{p(\omega\vert \mathcal{D}_\textrm{Expert},\Theta_\textrm{Expert})}p(\omega\vert \mathcal{D}_\textrm{Expert},\Theta_\textrm{Expert})dP(\Theta_\textrm{Expert}\vert \mathcal{D}_\textrm{Expert}),\\
		&=\int_{\Theta^{N_\textrm{Expert}}}\nabla_\omega  \log p(\omega\vert \mathcal{D}_\textrm{Expert},\Theta_\textrm{Expert})dP(\Theta_\textrm{Expert},\omega\vert \mathcal{D}_\textrm{Expert}),\\
		&=\int_{\Theta^{N_\textrm{Expert}}}\nabla_\omega  \log p(\omega\vert \mathcal{D}_\textrm{Expert},\Theta_\textrm{Expert})dP(\Theta_\textrm{Expert}\vert \omega, \mathcal{D}_\textrm{Expert})p(\omega\vert \mathcal{D}_\textrm{Expert}).
\end{align}
Substituting yields our desired result:
\begin{align}
	\nabla_\omega \log p(\omega \vert \mathcal{D}_\textrm{Expert})=\int_{\Theta^{N_\textrm{Expert}}}\nabla_\omega  \log p(\omega\vert \mathcal{D}_\textrm{Expert},\Theta_\textrm{Expert})dP(\Theta_\textrm{Expert}\vert \omega, \mathcal{D}_\textrm{Expert}).
\end{align}

Now, 
\begin{align}
	p(\omega\vert \mathcal{D}_\textrm{Expert},\Theta_\textrm{Expert})&=\frac{p(\mathcal{D}_\textrm{Expert}\vert \omega,\Theta_\textrm{Expert})p(\omega)}{\int p(\mathcal{D}_\textrm{Expert}\vert \omega,\Theta_\textrm{Expert})dP(\omega) },\\
	&=\frac{\prod_{i=1}^{N_\textrm{Expert}}p(\tau_i\vert \omega,\theta_i)p(\omega)}{\int\prod_{i=1}^{N_\textrm{Expert}}p(\tau_i\vert \omega,\theta_i)dP(\omega)},\\
	&=\frac{\prod_{i=1}^{N_\textrm{Expert}}\prod_{j=0}^{H_i-1}p(s_{j+1}\vert s_j,a_j,\theta_i)p(a_j\vert s_j,\omega,\theta_i)p(\omega)}{\int\prod_{i=1}^{N_\textrm{Expert}}\prod_{j=0}^{H_i-1}p(s_{j+1}\vert s_j,a_j,\theta_i)p(a_j\vert s_j,\omega,\theta_i)dP(\omega)},\\
	&=\frac{\prod_{i=1}^{N_\textrm{Expert}}\prod_{j=0}^{H_i-1}p(a_j\vert s_j,\omega,\theta_i)p(\omega)}{\int\prod_{i=1}^{N_\textrm{Expert}}\prod_{j=0}^{H_i-1}p(a_j\vert s_j,\omega,\theta_i)dP(\omega)},
\end{align}
where we have used the fact that each $p(s_{j+1}\vert s_j,a_j,\theta_i)$ has no dependence on $\omega$, and so will cancel in the fraction when deriving the final line. Now, substituting for the definition of the likelihood: 
\begin{align}
	p(\omega&\vert \mathcal{D}_\textrm{Expert},\Theta_\textrm{Expert})\\
	&\propto\underbrace{\exp\left(\sum_{i=1}^{N_\textrm{Expert}}\sum_{j=0}^{H_i-1}\left(\frac{\Psi^\star(s_j,a_j,\theta_i)^\top \omega}{\alpha}-\log z(s_j,\theta_i,\omega)\right)\right)}_{\textrm{Likelihood}}\underbrace{\exp\left(-\frac{\lVert \omega-\omega_0\rVert^2}{2\sigma_0^2}\right)}_{\textrm{Prior}},\\
	&=\exp\left(\sum_{i=1}^{N_\textrm{Expert}}\sum_{j=0}^{H_i-1}\left(\frac{\Psi^\star(s_j,a_j,\theta_i)^\top \omega}{\alpha}-\log z(s_j,\theta_i,\omega)\right)-\frac{\lVert \omega-\omega_0\rVert^2}{2\sigma_0^2}\right),
\end{align}
hence:
 \begin{align}
\nabla_\omega &\log	p(\omega\vert \mathcal{D}_\textrm{Expert},\Theta_\textrm{Expert})\\
&=\nabla_\omega\left( \sum_{i=1}^{N_\textrm{Expert}}\sum_{j=0}^{H_i-1}\left(\frac{\Psi^\star(s_j,a_j,\theta_i)^\top \omega}{\alpha}-\log z(s_j,\theta_i,\omega)\right)-\frac{\lVert \omega-\omega_0\rVert^2}{2\sigma_0^2}\right),\\
	\implies &g(\omega; \mathcal{D}_\textrm{Expert})\\
	=&\int_{\Theta^{N_\textrm{Expert}}}\nabla_\omega\Bigg( \sum_{i=1}^{N_\textrm{Expert}}\sum_{j=0}^{H_i-1}\left(\frac{\Psi^\star(s_j,a_j,\theta_i)^\top \omega}{\alpha}-\log z(s_j,\theta_i,\omega)\right)\\
	&\quad-\frac{\lVert \omega-\omega_0\rVert^2}{2\sigma_0^2}\Bigg)dP(\Theta_\textrm{Expert}\vert \omega, \mathcal{D}_\textrm{Expert}),\\
	=&\sum_{i=1}^{N_\textrm{Expert}}\mathbb{E}_{\theta_i \sim p(\theta_i\vert \omega,\tau_i)}\left[\nabla_\omega\left( \sum_{j=0}^{H_i-1}\left(\frac{\Psi^\star(s_j,a_j,\theta_i)^\top \omega}{\alpha}-\log z(s_j,\theta_i,\omega)\right)-\frac{\lVert \omega-\omega_0\rVert^2}{2\sigma_0^2}\right)\right].\label{eq:gradient}
\end{align}
Now, applying \Cref{proof:normalisation_gradient} and multiplying by $\alpha$ yields:
\begin{align}
	\alpha g(\omega; \mathcal{D}_\textrm{Expert})=&\sum_{i=1}^{N_\textrm{Expert}}\sum_{j=0}^{H_i-1}\mathbb{E}_{\theta_i \sim p(\theta_i\vert \omega,\mathcal{D}_\textrm{Expert})}\left[ \left(\Psi^\star(s_j,a_j,\theta_i)-\mathbb{E}_{a_i\sim p(a_i\vert s_j,\omega,\theta_i)}\left[ \Psi^\star(s_j,a_i,\theta_i) \right]\right)\right]\\
	&\quad-\frac{(\omega-\omega_0)}{\varsigma_0^2},
\end{align}
as required
\end{proof}
\end{theorem}

\renewcommand{\thesection}{\Alph{section}}
\setcounter{section}{6}
\subsection{Derivation of BRL Objective}
\label{app:BRL_objective_derivation}
Starting from the Bayesian RL objective:
\begin{align}
	J^\pi_\textrm{Bayes}&=\mathbb{E}_{\tau_\infty\sim p^\pi_\infty(\tau_\infty)}\left[\sum_{i=0}^\infty \gamma^i r_i\right],\\
	&=\mathbb{E}_{h_\infty\sim p^\pi_\infty(h_\infty)}\left[\sum_{i=0}^\infty \gamma^i \mathbb{E}_{r_i\sim p(r_i\vert h_i,a_i) }\left[r_i\right]\right].
\end{align}
Now, 
\begin{align}
	\mathbb{E}_{r_i\sim p(r_i\vert h_i,a_i) }\left[r_i\right]&=\mathbb{E}_{r_i\sim p^\textrm{IRL+COE}_\textrm{Bayes}(r_i\vert s_i,a_i) }\left[r_i\right],\\
	&=\begin{cases}
		\mathbb{E}_{k\sim p(k)}\left[\mathbb{E}_{r_i\sim p(r_i\vert s_i,a_i,k)}\left[r_i\right]\right], &s,a\in\mathcal{S}_\textrm{COE}\times \mathcal{A}_\textrm{COE},\\
		\mathbb{E}_{\omega\sim p(\omega)}\left[\mathbb{E}_{r_i\sim p(r_i\vert s_i,a_i,\omega)}\left[r_i\right]\right], &\textrm{otherwise},
	\end{cases}\\
	&=\begin{cases}
		\mathbb{E}_{k\sim p(k)}\left[k r_\textrm{max}\right], &s,a\in\mathcal{S}_\textrm{COE}\times \mathcal{A}_\textrm{COE},\\
		\mathbb{E}_{\omega\sim p(\omega)}\left[ \nu(s_i,a_i)^\top \omega\right], &\textrm{otherwise},
	\end{cases}\\
	&=\begin{cases}
		k^\star r_\textrm{max}, &s,a\in\mathcal{S}_\textrm{COE}\times \mathcal{A}_\textrm{COE},\\
	\nu(s_i,a_i)^\top \omega^\star_\textrm{Bayes}, &\textrm{otherwise},
	\end{cases}\\
	&=r^\textrm{IRL+COE}_\textrm{Bayes}(s_i,a_i),
\end{align}
hence:
\begin{align}
		J^\pi_\textrm{Bayes}&=\mathbb{E}_{h_\infty\sim p^\pi_\infty(h_\infty)}\left[\sum_{i=0}^\infty \gamma^i r^\textrm{IRL+COE}_\textrm{Bayes}(s_i,a_i)\right],\\
		&=\mathbb{E}_{\theta_\textrm{test}\sim p(\theta_\textrm{test})}\left[\mathbb{E}_{h_\infty\sim p^\pi_\infty(h_\infty\vert \theta)}\left[\sum_{i=0}^\infty \gamma^i r^\textrm{IRL+COE}_\textrm{Bayes}(s_i,a_i)\right]\right],
\end{align}
as required.

\section{The Role of Temperature in IRL}
\label{app:balancing_prior}

\begin{wrapfigure}{l}{0.3\textwidth}
\centering
\vspace{5mm}
\includegraphics[width=0.3\textwidth]{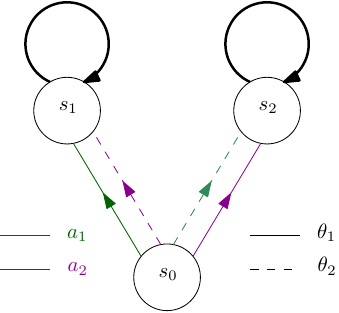}
\caption{Space of Three State MDPs}
\label{fig:prior_example}
\vspace{-5mm}
\end{wrapfigure}

We now provide an intuitive example to demonstrate how the prior influences the expert data in our Bayesian formulation. Consider the space of three state MDPs in \Cref{fig:prior_example}. The agent has a set of two possible actions $\mathcal{A}=\{a_1,a_2\}$. For $\theta=\theta_1$, the agent transitions to state $s_1$ deterministically after selecting $a_1$ or $s_2$ after selecting $a_2$. For $\theta=\theta_2$, the actions are reversed. In both MDPs, the initial state is $s_0$ and state $s_1$ or $s_2$ are terminal. The reward function depends only on states and is $r(s_0)=0$, $r(s_1)=1$, $r(s_2)=-1$. Let $\mathbb{I}(s)\in \{0,1\}^3$ denote the indicator feature vector where $I(s_n)$ is a one-hot vector where the $n$th element is 1, e.g.  $\mathbb{I}(s_1)= (0,1,0)^\top$. We represent this reward function using the linear form $r(s)=\mathbb{I}(s)^\top \omega^\star$ where $\omega^\star=(0,1,-1)^\top$. Finally, MDPs are allocated using a uniform prior. 

Expert data will always consist of trajectories that transition from state $s_0$ to state $s_1$. For our feature vector, we can derive the corresponding expert successor feature representation analytically: 
\begin{gather}
	\Psi^\star(s_0,a_1,\theta_1)=\mathbb{I}(s_0)+\frac{\gamma}{1-\gamma}\mathbb{I}(s_1),\ \Psi^\star(s_0,a_2,\theta_1)=\mathbb{I}(s_0)+\frac{\gamma}{1-\gamma}\mathbb{I}(s_2),\\
	\Psi^\star(s_0,a_2,\theta_2)=\mathbb{I}(s_0)+\frac{\gamma}{1-\gamma}\mathbb{I}(s_1),\ \Psi^\star(s_0,a_1,\theta_2)=\mathbb{I}(s_0)+\frac{\gamma}{1-\gamma}\mathbb{I}(s_2),\\
	 \Psi^\star(s_1,\cdot,\cdot)=\frac{1}{1-\gamma}\mathbb{I}(s_1),\  \Psi^
	 \star(s_2,\cdot,\cdot)=\frac{1}{1-\gamma}\mathbb{I}(s_2).
\end{gather}
Using the gradient update in \Cref{eq:stochastic_updates}, we see that updates in state $s_0$ will initially draw actions equally from $p(a_i\vert s_0,\omega,\cdot)$. This yields an initial gradient update of:
\begin{align}
	g_\omega^0&=\omega_0+ \eta_\omega^0  \frac{N_\textrm{Expert}{H_i}}{2} \left(\Psi^\star(s_0,a_1,\theta_1)-\frac{1}{2}\left[ \Psi^\star(s_0,a_1,\theta_1)+\Psi^\star(s_0,a_2,\theta_1) \right]\right)+\\
	  &\quad\quad\eta_\omega  \frac{N_\textrm{Expert}{H_i}}{2} \left(\Psi^\star(s_0,a_2,\theta_2)-\frac{1}{2}\left[ \Psi^\star(s_0,a_2,\theta_2)+\Psi^\star(s_0,a_1,\theta_2) \right]\right),\\
	  &=  \omega_0+\frac{\gamma\eta_\omega^0 N_\textrm{Expert}H_i}{2(1-\gamma)}\left[\mathbb{I}(s_1)-\mathbb{I}(s_2)\right].
\end{align}
For exposition, assume that there is no prior preference between $\omega_0^1$ and  $\omega_0^2$, and that $\omega_0^1=\omega_0^2=0$. We see that the initial gradient will update these values to:
\begin{align}
	\omega_1^1 = \frac{\gamma\eta_\omega^0 N_\textrm{Expert}H_i}{2(1-\gamma)},\quad \omega_1^2 = -\frac{\gamma\eta_\omega^0 N_\textrm{Expert}H_i}{2(1-\gamma)}.
\end{align}
We now consider the regime where the temperature parameter tends to zero $\alpha\rightarrow0$. For all future updates, as $\omega_1>\omega_2$, the model expert policy will tend towards a deterministic function that picks the action leading to state $s_1$: $p(a\vert s_0,\omega,\theta_i)=\delta(a=a_i)$. This means that all future updates $k\ge1$ from state $s_0$ will pull the reward parametrization back to the prior value:
\begin{align}
	g_\omega^k=\omega_k - \frac{\eta_\omega^k (\omega_k-\omega_0)}{\varsigma_0^2}.
\end{align}

In the regime where the temperature parameter tends to infinity $\alpha\rightarrow\infty$, the model expert policy will remain uniform over all actions. Under this assumption, all future updates $k\ge1$ from state $s_0$ will continue to increase the value of $\omega_1$ and decrease the value of $\omega_2$, whilst pulling $\omega_k$ back towards the prior in accordance with $\varsigma_0^2$. 
\begin{align}
	g_\omega^k &=  \omega_k+\frac{\gamma\eta_\omega^k N_\textrm{Expert}H_i}{2(1-\gamma)}\left[\mathbb{I}(s_1)-\mathbb{I}(s_2)\right]- \frac{\eta_\omega^k (\omega_k-\omega_0)}{\varsigma_0^2}.
\end{align}
When used in practice, we select $\alpha$ to be between $0$ and $\infty$. Our example reveals that the smaller the temperature parameter, the less the updates will separate values of reward for states that the expert visited vs that the expert could have visited. Conversely, when $\alpha$ is very large, this difference will grow and can only be counteracted by the prior variance $\varsigma_0$.

\section{Experiments}
\subsection{Computer resources}
The experiments were run on servers with eight recent NVIDIA GPUs (3080, A4500, or A5000).
The random seeds were run in parallel.
The experiments for the gridworld were the longest and took approximately an hour to run.

\subsection{Latent Inference Investigation Details}
\label{app:counterexample_experiment}
\begin{figure}[h]
    \centering
    \includegraphics[width=0.3\textwidth]{figures/counterexample_cmdp.png}
    \includegraphics[width=0.3\textwidth, trim={4em 4em 4em 4em}, clip]{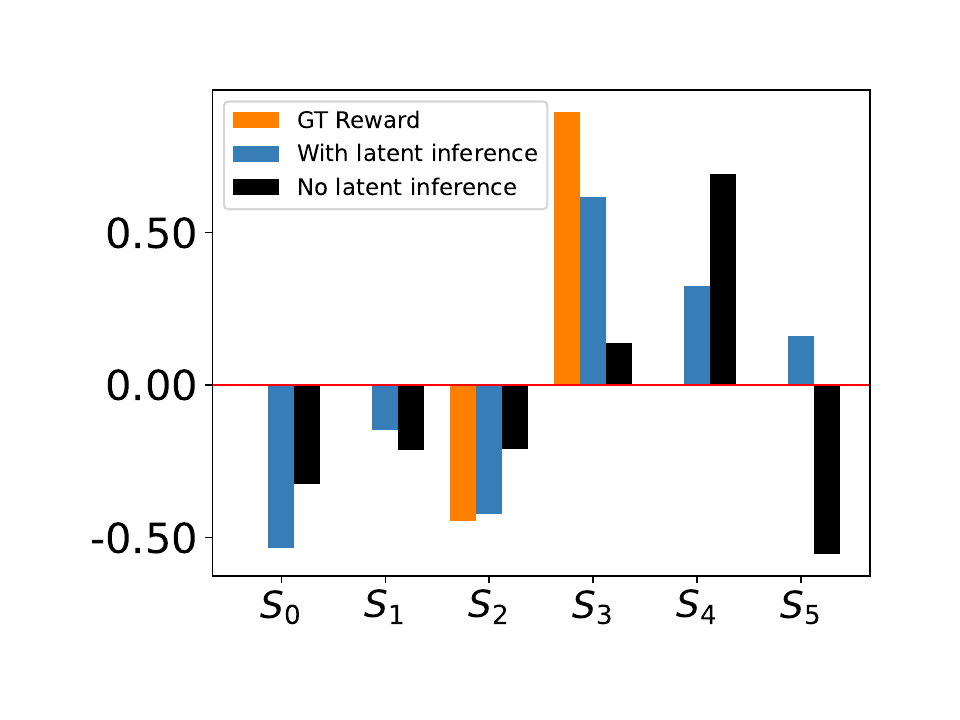}
    \caption{
    A demonstration of the necessity for latent inference.
    On the left, we repeat the visualization of the CMDP used in the experiments for convenience.
    On the right, we show the rewards learned with and without latent inference compared to the ground truth rewards.
    The reward vectors are normalized to unit norm.
    The y-axis is linearly scaled.
    }
    \label{fig:app_counterexample}
\end{figure}
\Cref{fig:app_counterexample} shows the learned rewards in the latent inference experiment described in \Cref{subsec:investigating-latent-inference}.
The learned rewards are somewhat hard to interpret because in addition to the scaling and shifting information being lost in IRL, there is no pressure for the algorithms to keep the reward vectors sparse.
Instead, they are only trying to optimize \Cref{eq:stochastic_updates}.
Nevertheless, the rewards learned with latent inference have the same ordering between the critical states $s_1$, $s_2$, and $s_3$ as the ground truth reward.
In contrast, the rewards learned with naive IRL do not differentiate between $s_1$ and $s_2$, which results in the learned policies not taking the path through $s_1$ even when it is available.

\subsection{Tiger Treasure Details}
The Tiger-Treasure problem is displayed in \Cref{fig:tiger_treasure} and described in \Cref{sec:counterexample}.
The agent starts in state $S_0$ and can either listen or open a door.
The listening action transitions the agent to a 'hint' state revealing with probability $p=0.85$ the location of the Tiger.
After the agent opens a door, it is either roared by the tiger or collects the treasure and arrives in the terminal state $S_T$.
For learning the contextual IRL reward, we use the hyperparameters presented in Table~\ref{tab:tt_irl_hyperparameters} with a default value of $\omega_0=(0,0,0,0,0,0)$ in the Laplace approximation Equation~\ref{eq:stochastic_updates}.
The reward prior is introduced after having rescaled the IRL reward between $r_\text{min}=-100$ and $r_\text{max}=10$. To enhance the training process, we normalize the rewards before training the DQN policies.
The hyperparameters presented in Table~\ref{tab:tt_dqn_hyperparameters}.

\begin{table}[H]
\centering
\begin{tabular}{l|l}
\hline
\textbf{Parameter} & \textbf{Value} \\
\hline
Number of parallel environments &  500 \\
Number of steps per rollout &  50 \\
Number of updates &  5000 \\
$\epsilon$-greedy $\epsilon$ &  0.5 \\
Maximum gradient norm &  0.5 \\
Discount $\gamma$ &  0.99 \\
$\alpha$ &  0.01 \\
$\varsigma^2_0$ &  100.0 \\
Target SF update rate &  50 \\
SF learning rate &  $1 \times 10^{-3}$ \\
Reward learning rate &  $1 \times 10^{-2}$ \\
Replay buffer size &  50000 \\
Batch size (trajectories) &  500 \\
$r_\text{max}$ &  10.0 \\
$r_\text{min}$ &  -100.0 \\
\hline
\end{tabular}
\caption{IRL Hyperparameters for the Tiger Treasure problem.}
\label{tab:tt_irl_hyperparameters}
\end{table}

\begin{table}[H]
\centering
\begin{tabular}{l|l}
\hline
\textbf{Parameter} & \textbf{Value} \\
\hline
Number of parallel environments & 16 \\
Number of steps per rollout & 50 \\
Total number of updates & 20000 \\
Learning rate & $1 \times 10^{-4}$ \\
Discount $\gamma$ & 0.99 \\
Target network update rate & 1 \\
Replay buffer size (number of full trajectories) & 200000 \\
Batch size (number of full trajectories) & 100 \\
Starting value for $\epsilon$ & 1.0 \\
Final value for $\epsilon$ & 0.05 \\
Fraction of updates after $\epsilon$ schedule finishes & 0.5 \\
\hline
\end{tabular}
\caption{DQN Hyperparameters for the Tiger Treasure problem.}
\label{tab:tt_dqn_hyperparameters}
\end{table}

\subsection{Tiger Treasure Maze Details}
\label{app:tiger_maze}
The agent starts in the middle of the two doors.
It can choose to move in the cardinal directions or listen.
After the agent opens a door, it will either collect a treasure or be roared at by a tiger for one timestep.
Then, depending on what happened, it gets transported to the nearest grid cell to the $Tiger$ or $Gold$.
The expert takes the shortest path to $Gold$ and never listens.
The state of the agent is defined as the $X-Y$ coordinates of the agent and an indicator variable, which indicates the result of the listening action.
After taking a listening action, the agent is transported to a state with the same $X-Y$ coordinates but with the indicator showing the true value of $\theta$.
The indicator states have the same dynamics as the corresponding normal states.
The coordinates and the indicator are encoded as one-hot vectors.
The reward feature $\nu$ is the full table of all states visitable by the agent.
Since the dynamics are deterministic given $\theta$, the inference model labels the trajectories with the true $\theta$ if it is revealed on the trajectory.

The hyperparameters used for BIG in the maze experiment are shown in Table~\ref{tab:maze_irl_hyperparameters}.
The hyperparameters for DQN in the maze experiment are shown in Table~\ref{tab:maze_dqn_hyperparameters}.
A recurrent neural network is used for the policy architecture.
A separate MLP network is used for implementing the critic.

\begin{table}[H]
\centering
\begin{tabular}{l|l}
\hline
\textbf{Parameter} & \textbf{Value} \\
\hline
Number of parallel environments & 16 \\
Number of steps per rollout & 40 \\
Number of updates & 20000 \\
$\epsilon$-greedy $\epsilon$ & 0.5 \\
Maximum gradient norm & 0.5 \\
Discount $\gamma$ & 0.99 \\
$\alpha$ & 0.01 \\
$\varsigma^2_0$ & 1.0 \\
Target SF update rate & 50 \\
SF learning rate & $3 \times 10^{-4}$ \\
Reward learning rate & $3 \times 10^{-3}$ \\
Replay buffer size (number of full trajectories) & 10000 \\
Batch size (number of full trajectories) & 100 \\
$k$ & 0.01 \\
$r_\text{max}$ & 1.0 \\
$r_\text{min}$ & -0.05 \\
\hline
\end{tabular}
\caption{IRL Hyperparameters for the Maze problem.}
\label{tab:maze_irl_hyperparameters}
\end{table}

\begin{table}[H]
\centering
\begin{tabular}{l|l}
\hline
\textbf{Parameter} & \textbf{Value} \\
\hline
Number of parallel environments & 16 \\
Number of steps per rollout & 40 \\
Total number of updates & 100000 \\
Learning rate & $1 \times 10^{-4}$ \\
Discount $\gamma$ & 0.99 \\
Target network update rate & 1 \\
Replay buffer size (number of full trajectories) & 10000 \\
Batch size (number of full trajectories) & 100 \\
Starting value for $\epsilon$ & 1.0 \\
Final value for $\epsilon$ & 0.05 \\
Fraction of updates after $\epsilon$ schedule finishes & 0.5 \\
\hline
\end{tabular}
\caption{DQN Hyperparameters for the Maze problem.}
\label{tab:maze_dqn_hyperparameters}
\end{table}